\documentclass[journal, 10pt,twocolumn]{IEEEtran}

\usepackage{amssymb}
\usepackage[all,arc]{xy}
\usepackage{mathrsfs}
\usepackage{algorithm2e}
\usepackage{amsbsy}
\usepackage{amsfonts}
\usepackage{amsmath}
\usepackage{amssymb}
\usepackage{amsthm}
\usepackage{array}
\usepackage{authblk}

\usepackage{bbm} 
\usepackage[font=footnotesize,labelfont=bf]{caption}
\usepackage{color}
\usepackage[mathscr]{eucal}
\usepackage{mathrsfs}

\usepackage{float}
\usepackage{hyperref} 
\usepackage{stmaryrd}
\usepackage[font=scriptsize,labelfont=scriptsize]{subcaption}
\usepackage{graphicx}
\usepackage{url}
\usepackage{verbatim}
\usepackage{adjustbox}
\usepackage[usenames,dvipsnames,svgnames,table]{xcolor}
\usepackage{multirow,array}

\makeatletter
\def\ps@pprintTitle{%
 \let\@oddhead\@empty
 \let\@evenhead\@empty
 \def\@oddfoot{}%
 \let\@evenfoot\@oddfoot}
\makeatother

\title{Spectral-Spatial Diffusion Geometry for Hyperspectral Image Clustering}

\author{James M. Murphy \and ~ Mauro Maggioni
\IEEEcompsocitemizethanks{
\IEEEcompsocthanksitem J.M. Murphy is with the Department of Mathematics at Tufts University; email: JM.Murphy@tufts.edu\\
\IEEEcompsocthanksitem M. Maggioni is with the Department of Mathematics and the Department of Applied Mathematics and Statistics at Johns Hopkins University; email: mauromaggionijhu@icloud.com
}
}

\begin{document}

\maketitle

\begin{abstract}

An unsupervised learning algorithm to cluster hyperspectral image (HSI) data is proposed that exploits spatially-regularized random walks.  Markov diffusions are defined on the space of HSI spectra with transitions constrained to near spatial neighbors.  The explicit incorporation of spatial regularity into the diffusion construction leads to smoother random processes that are more adapted for unsupervised machine learning than those based on spectra alone.  The regularized diffusion process is subsequently used to embed the high-dimensional HSI into a lower dimensional space through diffusion distances.  Cluster modes are computed using density estimation and diffusion distances, and all other points are labeled according to these modes.  The proposed method has low computational complexity and performs competitively against state-of-the-art HSI clustering algorithms on real data.  In particular, the proposed spatial regularization confers an empirical advantage over non-regularized methods. 
\end{abstract}

\section{Introduction}

As the volume of data captured by remote sensors grows unabated, human capacity for providing labeled training datasets is strained.  State-of-the-art supervised learning algorithms (e.g. deep neural networks) require very large training sets to fit the massive number of parameters associated to their high-complexity architectures.  In order to take advantage of the deluge of unlabeled remote sensing data available, new methods that are \emph{unsupervised}---requiring no training data---are necessary.  

This article proposes an efficient unsupervised clustering algorithm for hyperspectral imagery (HSI) that exploits not only low-dimensional geometry in the high-dimensional space of spectra, but also the spatial regularity in the 2-dimensional image structure of the pixels.  This is achieved by considering \emph{spectral-spatial diffusion geometry} as captured by spatially regularized diffusion distances \cite{Coifman2005, Coifman2006}.  These distances are integrated into the recently proposed \emph{diffusion learning} algorithm, which has achieved competitive performance versus benchmark and state-of-the-art unsupervised HSI clustering algorithms \cite{Murphy2018_Diffusion, Murphy2018_Unsupervised}.  

The remainder of this article is organized as follows.  Background on HSI clustering and diffusion geometry is presented in Sec. \ref{sec:Background}.  The proposed algorithm is described and evaluated in Sec. \ref{sec:Algorithm} and \ref{sec:Results}, respectively.  Conclusions and future research directions are presented in Sec. \ref{sec:Conclusions}.

\section{Background}\label{sec:Background}

The unsupervised clustering of HSI $X=\{x_{n}\}_{n=1}^{N}\subset\mathbb{R}^{D}$---understood as a point cloud---consists in providing labels $\{y_{n}\}_{n=1}^{N}$ to each data point without access to labeled training data.  The total number of pixels in the image is $N$, and the number of spectral bands is $D$.  Typically $D$ is quite large, causing traditional statistical methods to perform sub-optimally.  However, the different clusters in the data---corresponding to regions with distinct material properties---often exhibit low-dimensional, though nonlinear, structure.  In order to efficiently exploit this structure, methods for clustering HSI that learn the underlying nonlinear geometry have been developed, including methods based on non-negative matrix factorization \cite{Gillis2015}, regularized graph Laplacians \cite{Meng2017}, angle distances \cite{Erturk2006_Unsupervised}, and deep neural networks \cite{Tao2015_Unsupervised}.

Recently, the \emph{diffusion geometry}  \cite{Coifman2005,Coifman2006} of $X$ has been proposed in order to infer the latent clusters.  The \emph{diffusion distance at time $t$} between $x,y\in X$, denoted $d_{t}(x,y)$, is a notion of distance determined by the underlying geometry of the point cloud $X$.  The computation of $d_{t}$ involves constructing a weighted, undirected graph $\mathcal{G}$ with vertices corresponding to the $N$ points in $X$, and weighted edges stored in the $N\times N$ weight matrix $W(x,y):=\exp(-\|x-y\|_{2}^{2}/\sigma^{2})$ if $x\in NN_{k}(y), W(x,y)=0$ otherwise for some scaling parameter $\sigma$ and with $NN_{k}(x)$ the set of $k$-nearest neighbors of $y$ in $X$ with respect to Euclidean distance.  Typically $\sigma$ is chose adaptively, for example as some multiple of average distance to nearest neighbors \cite{Zelnik2005_Self}, and $k\ll n$ is chosen so that $W$ is sparse; we set $k=100$ in all experiments.  Let $P(x,y)={W(x,y)}\big/{\deg(x)}$ be a Markov diffusion matrix defined on $X$, where $\deg(x):=\sum_{y\in X}W(x,y)$ is the degree of $x$.  The \emph{diffusion distance at time $t$} is 
\begin{equation}
d_{t}(x,y):=\sqrt{\sum\nolimits_{u\in X} (P^{t}(x,u)- P^{t}(y,u))^{2}}.\label{e:diffdist}
\end{equation}The computation of $d_{t}(x,y)$ involves summing over all paths of length $t$ connecting $x$ to $y$, so $d_{t}(x,y)$ is small if $x,y$ are similar according to $P^{t}$.  

The eigendecomposition of $P$ yields fast algorithms to compute $d_{t}$: the matrix $ P$ admits a spectral decomposition (under mild conditions, see \cite{Coifman2006}) with eigenvectors $\{\Phi_{n}\}_{n=1}^{N}$ and eigenvalues $\{\lambda_{n}\}_{n=1}^{N}$, where $1=\lambda_{1}\ge |\lambda_{2}|\ge \dots\ge|\lambda_{N}|$.  The diffusion distance \eqref{e:diffdist} can then be written as
\begin{align}\label{eqn:DD_eigen}d_{t}(x,y)=\sqrt{\sum\nolimits_{n=1}^{N}\lambda_{n}^{2t}(\Phi_{n}(x)-\Phi_{n}(y))^{2}}\,.\end{align}  
The weighted eigenvectors $\{\lambda_{n}^{t}\Phi_{n}\}_{n=1}^{N}$ are new data-dependent coordinates of $X$, which are nearly geometrically intrinsic \cite{Coifman2005}.  The parameter $t$ measures how long the diffusion process on $\mathcal{G}$ has run: small values of $t$ allow a small amount of diffusion, which may prevent the interesting geometry of $X$ from being discovered.  On the other hand, large $t$ allow the diffusion process to run for so long that the fine geometry is homogenized.  In all our experiments we set $t=30$; see \cite{Murphy2018_Unsupervised} and \cite{Maggioni2018_Clustering} for empirical and theoretical analyses of $t$, respectively.

If the underlying graph $\mathcal{G}$ is connected, $|\lambda_n|< 1$ for $n>1$, so that the sum (\ref{eqn:DD_eigen}) may approximated by its truncation at some suitable $2\le M\ll N$.  In our experiments, $M$ was set to be the value at which the decay of the eigenvalues $\{\lambda_{n}\}_{n=1}^{N}$ begins to taper \cite{Murphy2018_Unsupervised}. The subset $\{\lambda_{n}^{t}\Phi_{n}\}_{n=1}^{M}$ used in the computation of $d_{t}$ is a dimension-reduced set of diffusion coordinates.  Indeed, the mapping \begin{align}\label{eqn:DiffusionMapsDR}x\mapsto (\lambda_{1}^{t}\Phi_{1}(x),\lambda_{2}^{t}\Phi_{2}(x),\dots,\lambda_{M}^{t}\Phi_{M}(x))\end{align} may be understood as a form of dimension reduction from the ambient space $\mathbb{R}^{D}$ to $\mathbb{R}^{M}$.  The truncation also enables us to compute only the first $M=O(1)$ eigenvectors, reducing computational complexity.  

Diffusion maps consider the data $X$ as a point cloud in $\mathbb{R}^{D}$.  If the data has organizing structure beyond its $D$-dimensional coordinates, this information can be incorporated into the diffusion maps construction by modifying the underlying transition matrix $P$.  In the case of HSI, each point is not only a high dimensional spectra, but also a pixel arranged in an image.  In particular, the HSI enjoys \emph{spatial regularity}, in the sense that points in a particular class are likely to have their nearest spatial neighbors being in the same class.  The incorporation of spatial information into supervised learning algorithms is known to improve empirical performance for a variety of data sets and methods \cite{Fauvel2012spatial, Cahill2014_Schroedinger,Zhang2016_Spectral}.  In this article, we propose to extend the recently proposed \emph{diffusion learning} unsupervised clustering framework \cite{Murphy2018_Unsupervised} by directly incorporating spatial information into the underlying diffusion matrix $P$.  

\section{Description of Algorithm}\label{sec:Algorithm}

\begin{figure}[!htb]
\centering
\includegraphics[width=.24\textwidth]{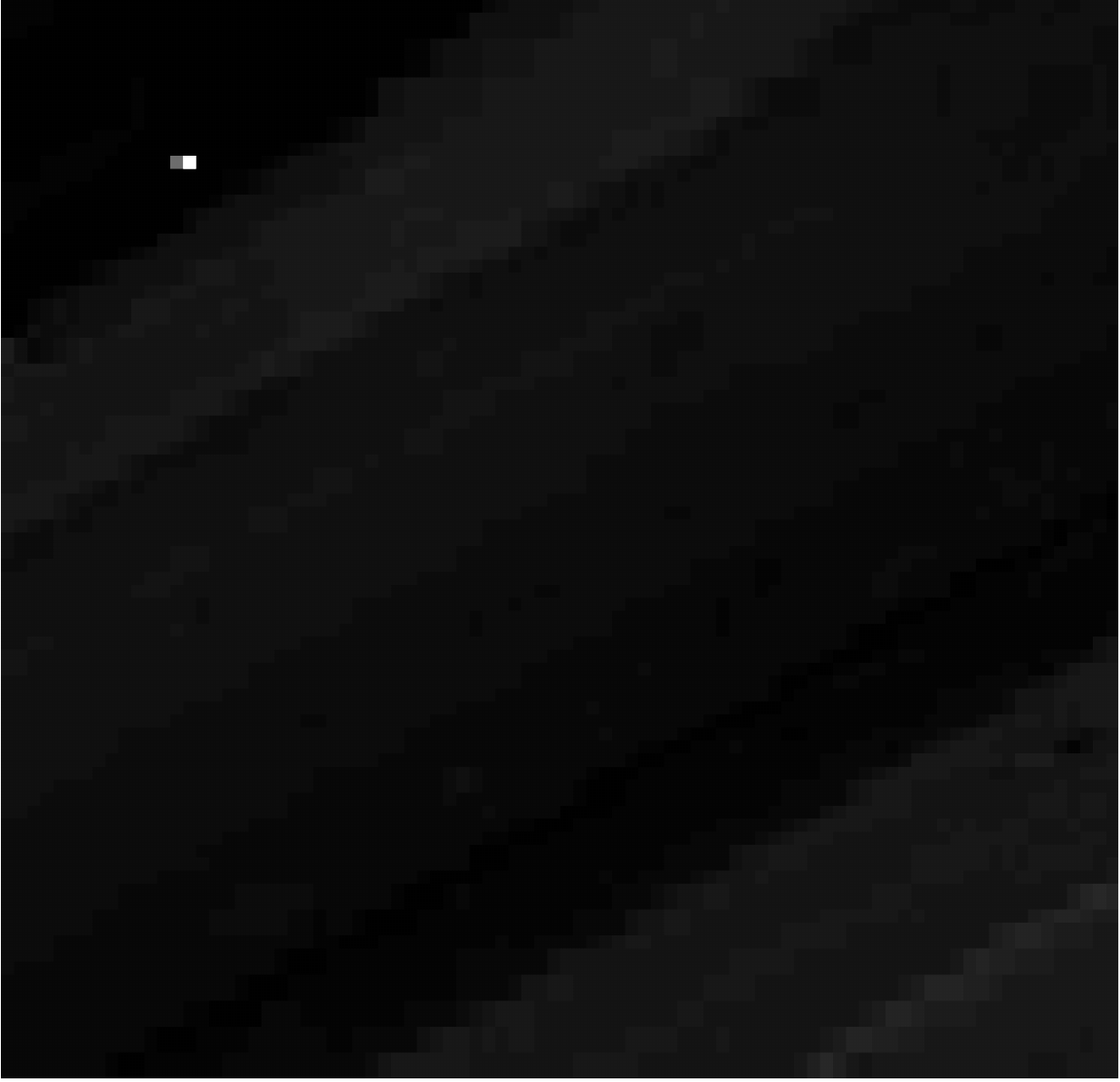}
\includegraphics[width=.24\textwidth]{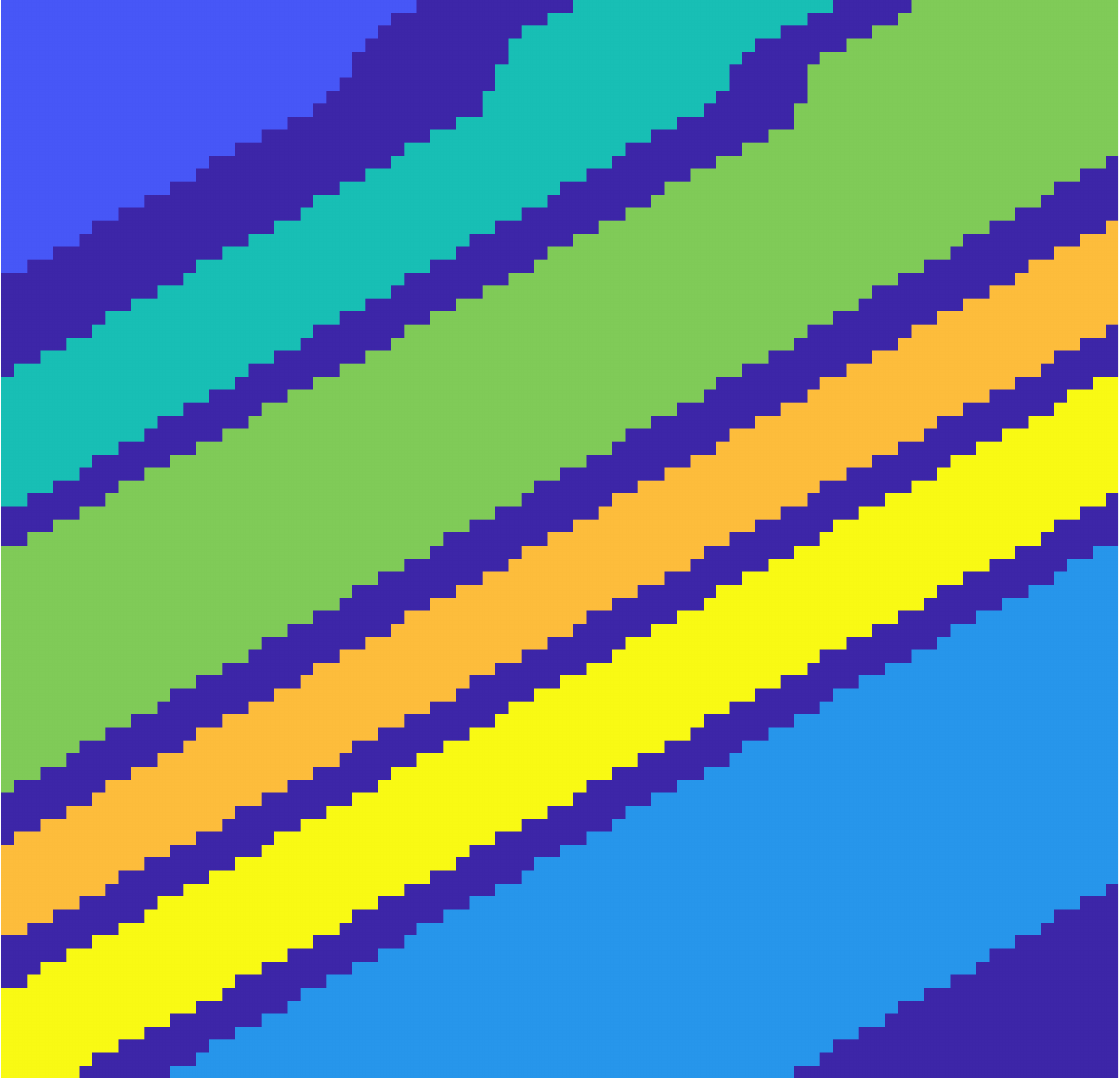}
\includegraphics[width=.24\textwidth]{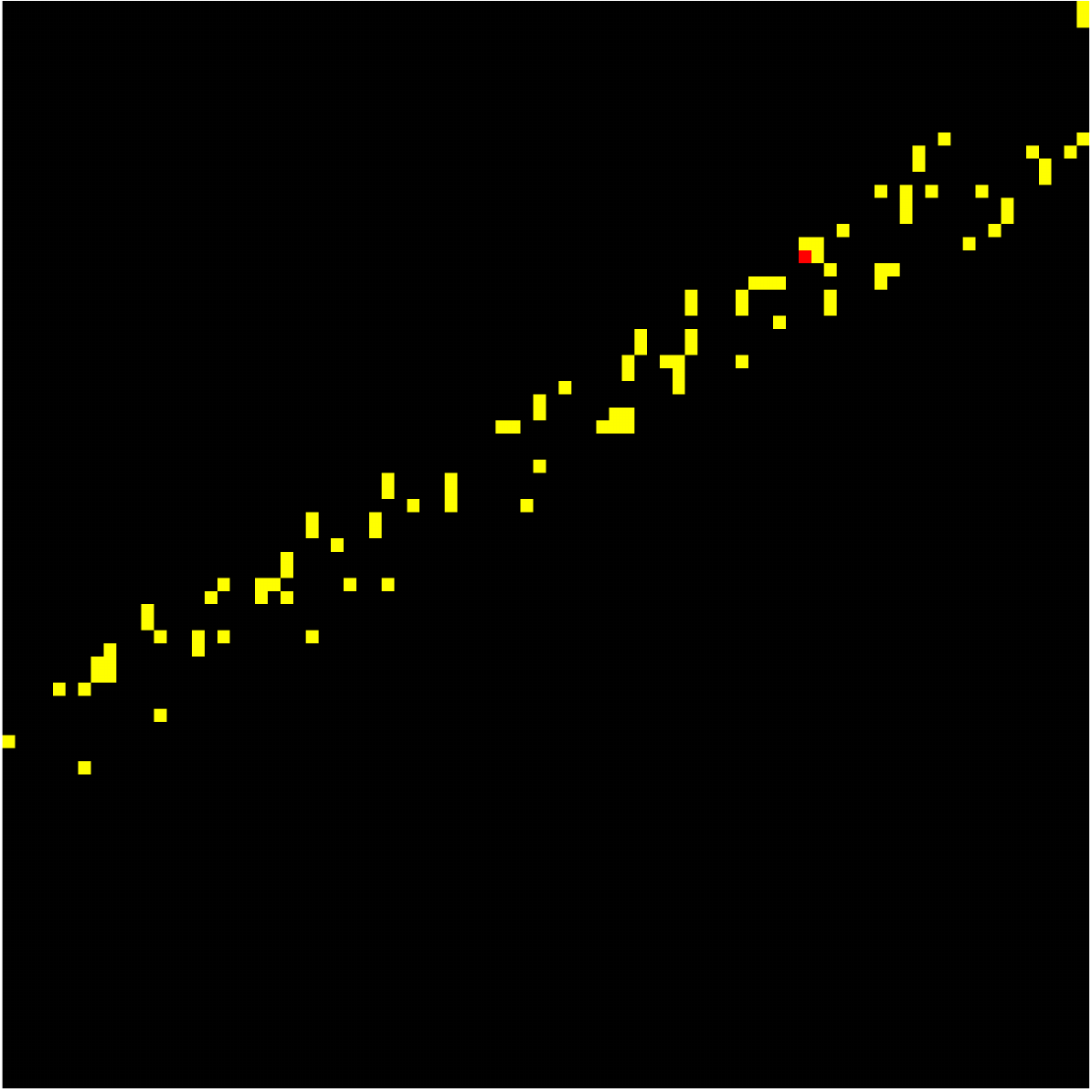}
\includegraphics[width=.24\textwidth]{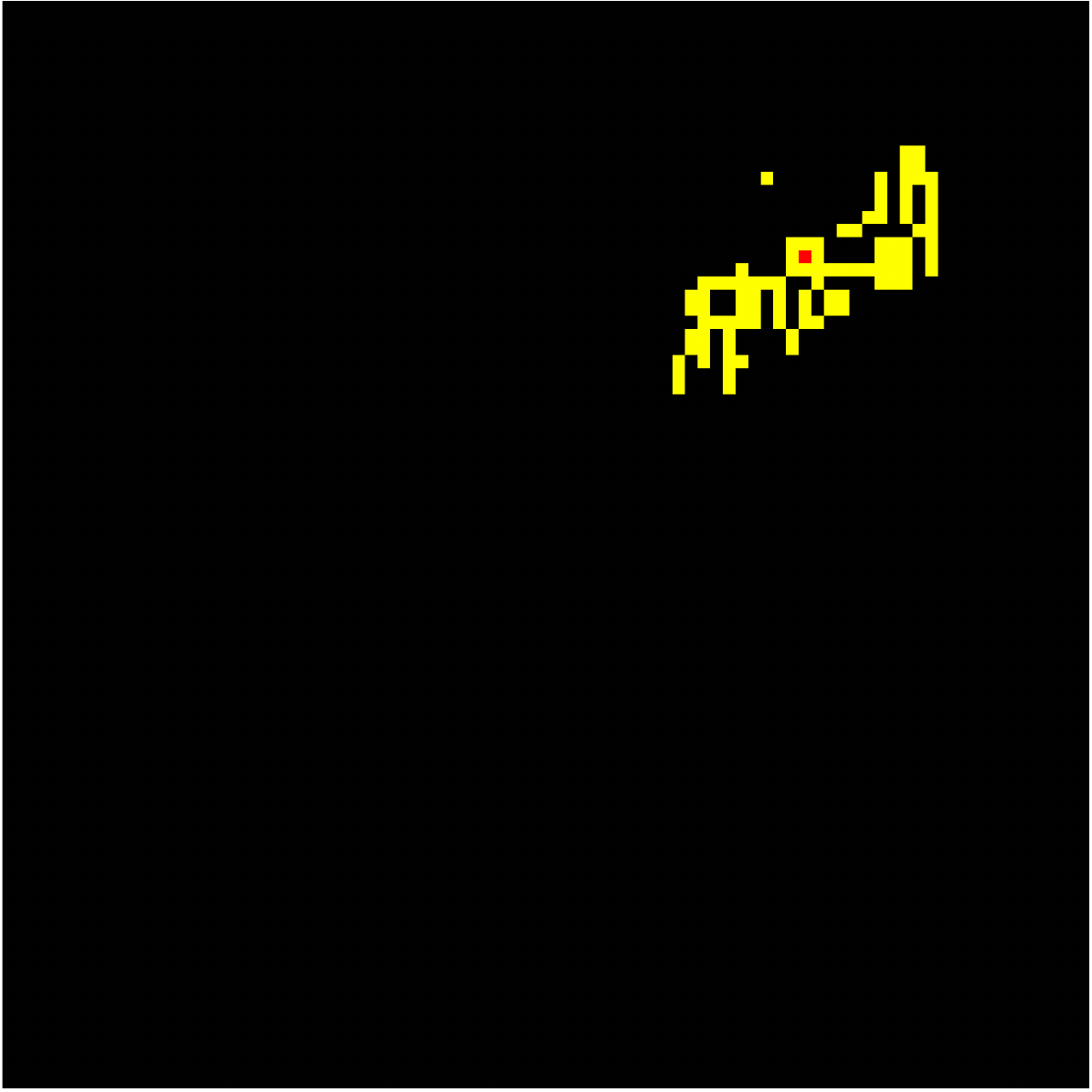}
\caption{\label{fig:SpatialNeighbors}The Salinas A data set is $86\times 83$ and contains 6 classes, all of which are well-localized spatially.  The dataset was captured over Salinas Valley, CA, by the AVRIS sensor.  The spatial resolution is 3.7 m/pixel.  The image contains 224 spectral bands.  \emph{Top left:} projection of the HSI onto its first principal component.  \emph{Top right}: ground truth (GT).  \emph{Bottom left:}  100 nearest neighbors (yellow) of a pixel (red) without spatial regularization.  \emph{Bottom right:}  100 nearest neighbors (yellow) of a pixel (red) with spatial regularization.} 
\end{figure}

The proposed algorithm first constructs a Markov diffusion matrix, $P$, under the constraint that pixels may only be connected to other pixels that are within some spatial radius $r$.  Figure \ref{fig:SpatialNeighbors} shows how nearest neighbors with and without this spatial constraint differ.  The construction of $P$ and the corresponding eigenpairs used to compute the diffusion distances are described in Algorithm \ref{alg:SSDM}.

\RestyleAlgo{algoruled}
\LinesNumbered
\begin{algorithm}[h]
\SetAlCapSkip{5em}	\caption{Spectral-Spatial Diffusion Maps}
	 \label{alg:SSDM}
  	\emph{Input}: $X, r$.\\
  	Connect each element $x\in X$ to its $k=100$ nearest neighbors within spatial radius $r$, call them $y$, with weight $\exp(-\|x-y\|_{2}^{2}/\sigma^{2})$.\\
  	Let $D$ be the diagonal degree matrix such that $D_{ii}=\sum_{j=1}^{n}W_{ij}$, and set $P=D^{-1}W$.\\
  	Compute the top $M$ eigenpairs of $P$, $\{(\lambda_{n},\Phi_{n})\}_{n=1}^{M}$.\\
  	\emph{Output:} $\{(\lambda_{n},\Phi_{n})\}_{n=1}^{M}$.
\end{algorithm}

Once the eigenpairs are computed, pairwise diffusion distances are simply Euclidean distances in the new coordinate system (\ref{eqn:DiffusionMapsDR}).  Cluster modes are computed as points maximizing $\mathcal{D}_{t}=p(x)\rho_{t}(x)$, where $p(x)$ is a kernel density estimator and $\rho_{t}(x)$ is the diffusion distance of a point to its nearest neighbor of higher density.  The mode detection algorithm is summarized in Algorithm \ref{alg:modes}; see \cite{Murphy2018_Unsupervised} for details.

\RestyleAlgo{algoruled}
\LinesNumbered
\begin{algorithm}[h]
\SetAlCapSkip{5em}	\caption{Geometric Mode Detection Algorithm}
	 \label{alg:modes}
  	\emph{Input}: $X, K, t$.\\
  	Compute a kernel density estimate $p(x_{n})$ for each $x_n\in X$.\\
	Compute diffusion distances using Algorithm \ref{alg:SSDM} and (\ref{eqn:DD_eigen}).\\
  	Compute $\{\rho_{t}(x_{n})\}_{n=1}^{N}$, the diffusion distance from each point to its nearest neighbor in diffusion distance of higher empirical density.\\
  	Compute the learned modes $\{x_{i}^{*}\}_{i=1}^{K}$ as the $K$ maximizers of $\mathcal{D}_{t}(x_{n})=p(x_{n})\rho_{t}(x_{n})$.\\
  	\emph{Output:} $\{x_{i}^{*}\}_{i=1}^{K}, \{p(x_{n})\}_{n=1}^{N}, \{\rho_{t}(x_{n})\}_{n=1}^{N}$.
\end{algorithm}

From the modes, points are labeled iteratively---from highest density to lowest density---according to their nearest spectral neighbor of higher density that has already been labeled, unless it is the case that such a labeling would strongly violate spatial regularity.  In that case, points are labeled according to their spatial nearest neighbors.  The spectral-spatial labeling scheme is summarized in Algorithm \ref{alg:labels}, and its crucial parameters and the role of the labeling spatial regularization are discussed at length in \cite{Murphy2018_Unsupervised}.  

\RestyleAlgo{algoruled}
\LinesNumbered
\begin{algorithm}[h]
\SetAlCapSkip{5em}
  \caption{Labeling Algorithm\label{alg:labels}}
  \emph{Input:} $\{x_{i}^{*}\}_{i=1}^{K}, \{p(x_{n})\}_{n=1}^{N}$, $\{\rho_{t}(x_{n})\}_{n=1}^{N}$.\\
  Assign each mode a unique label.\\
  \emph{Stage 1}: Iterating through the remaining unlabeled points in order of decreasing density, assign each point the same label as its nearest spectral neighbor (in diffusion distance) of higher density, unless the spatial consensus label exists and differs, in which case the point is left unlabeled.\\
  \emph{Stage 2}: Iterating in order of decreasing density among unlabeled points, assign each point the consensus spatial label, if it exists, otherwise the same label as its nearest spectral neighbor of higher density.  \\
  \emph{Output:} Labels $\{y_{n}\}_{n=1}^{N}$.
  
\end{algorithm}

The proposed method---consisting of Algorithms \ref{alg:SSDM}, \ref{alg:modes}, and \ref{alg:labels}---is called \emph{spatially-regularized diffusion learning (SRDL)}.  

\section{Experimental Results}\label{sec:Results}

We evaluate the SRDL algorithm on $3$ real HSI datasets.  The Indian Pines, Salinas A, and Kennedy Space Center datasets considered are standard, have ground truth, and are publicly available\footnote{\url{http://www.ehu.eus/ccwintco/index.php?title=Hyperspectral_Remote_Sensing_Scenes}}.  In order to quantitatively evaluate the unsupervised results in the presence of ground truth (GT), we consider three metrics.  \emph{Overall accuracy (OA)} is total number of correctly labeled pixels divided by the total number of pixels, which values large classes more than small classes.  \emph{Average accuracy (AA)} is the average, over classes, of the OA of each class, which values small classes and large classes equally.  \emph{Cohen's }$\kappa$-\emph{statistic} ($\kappa$) measures agreement across two labelings adjusted for random chance \cite{Banerjee1999}.  

We note that the Indian Pines and Kennedy Space Center datasets are restricted to subsets in the spatial domain, due to well-documented challenges of unsupervised methods for data containing a large number of classes \cite{Zhu2017}.  These datasets are restricted to reduce the number of classes and achieve meaningful clusters.  The Salinas A dataset is considered in its entirety.  We remark that results on small subsets can be patched together \cite{Murphy2018_Unsupervised}; such results are not shown here for reasons of space.  While the proposed method automatically estimates the number of clusters based on the decay of $\mathcal{D}_{t}$, the number of class labels in the ground truth images were used as parameter $K$ for all clustering algorithms to make a fair comparison with methods that cannot reliably estimate the number of clusters. 

Since the proposed and comparison methods are unsupervised, experiments are performed on the entire dataset, including points without ground truth labels.  The labels for pixels without ground truth are not accounted for in the quantitative evaluation of the algorithms tested.

\subsection{Comparison Methods}
\label{s:comparisonmethods}

We consider 13 benchmark and state-of-the-art methods of HSI clustering for comparison.  The benchmark methods are: \emph{$K$-means} \cite{Friedman2001} run directly on $X$;  {\em{principal component analysis} (PCA)}  followed by $K$-means;  {\em{independent component analysis} (ICA)} followed by $K$-means\cite{Comon1994, Hyvarinen1999, Hyvarinen2000}\footnote{\url{https://www.cs.helsinki.fi/u/ahyvarin/papers/fastica.shtml}}; {\em Gaussian random projections} followed by $K$-means \cite{Dasgupta2000}; \emph{DBSCAN} \cite{Ester1996}; \emph{spectral clustering} (SC) \cite{Ng2001}; and {\em{Gaussian mixture models}} (GMM) \cite{Acito2003}, with parameters determined by expectation maximization.  

The recent, state-of-the-art clustering methods considered are: \emph{sparse manifold clustering and embedding (SMCE)} \cite{Elhamifar2011, Elhamifar2013}\footnote{\url{http://vision.jhu.edu/code/}}, which fits the data to low-dimensional, sparse structures, and then applies spectral clustering; \emph{hierarchical clustering with non-negative matrix factorization (HNMF)} \cite{Gillis2015}\footnote{\url{https://sites.google.com/site/nicolasgillis/code}}, which has shown excellent performance for HSI clustering when the clusters are generated from a single endmember; a graph-based method based on the Mumford-Shah segmentation \cite{Mumford1989}\cite{Meng2017}, related to spectral clustering, and called \emph{fast Mumford-Shah (FMS)} in this article\footnote{\url{http://www.ipol.im/pub/art/2017/204/?utm_source=doi}}; \emph{fast search and find of density peaks clustering} (FSFDPC) algorithm  \cite{Rodriguez2014}, which has been shown effective in clustering a variety of data sets; and two variants of the recently proposed \emph{diffusion learning} algorithm, in which the labeling process considers only spectral information (DL) or spectral and spatial information (DLSS) \cite{Murphy2018_Unsupervised}.  

Among the comparison methods, the proposed method bears closest resemblance to the DL and DLSS methods.    SRDL and these two methods differ primarily in how the underlying geometry for clustering is learned.  In DL and DLSS, the diffusion geometry is computed though $P$ by considering the HSI only as a spectral point cloud. SRDL regularizes the construction of $P$ by incorporating spatial information into the nearest neighbors construction.  The proposed SRDL method also bears similarity to SC, SMCE, and FMS since all these methods use data-driven graphs.  The FSFDPC algorithm uses a mode detection scheme similar to SRDL, but without diffusion geometry or spatial information.  

\subsection{Indian Pines Data}

\begin{figure}[!htb]
\centering
\includegraphics[width=.24\textwidth]{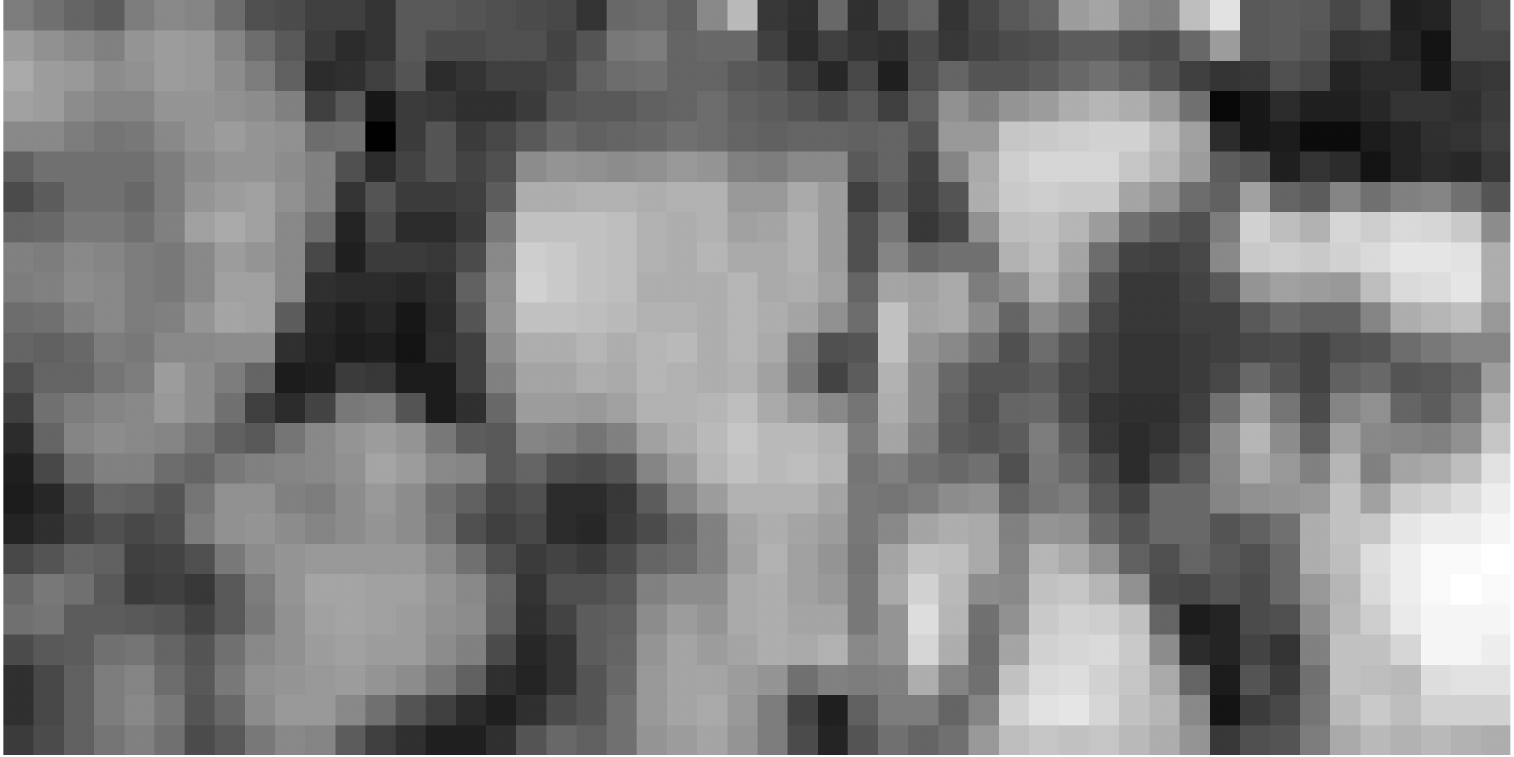}
\includegraphics[width=.24\textwidth]{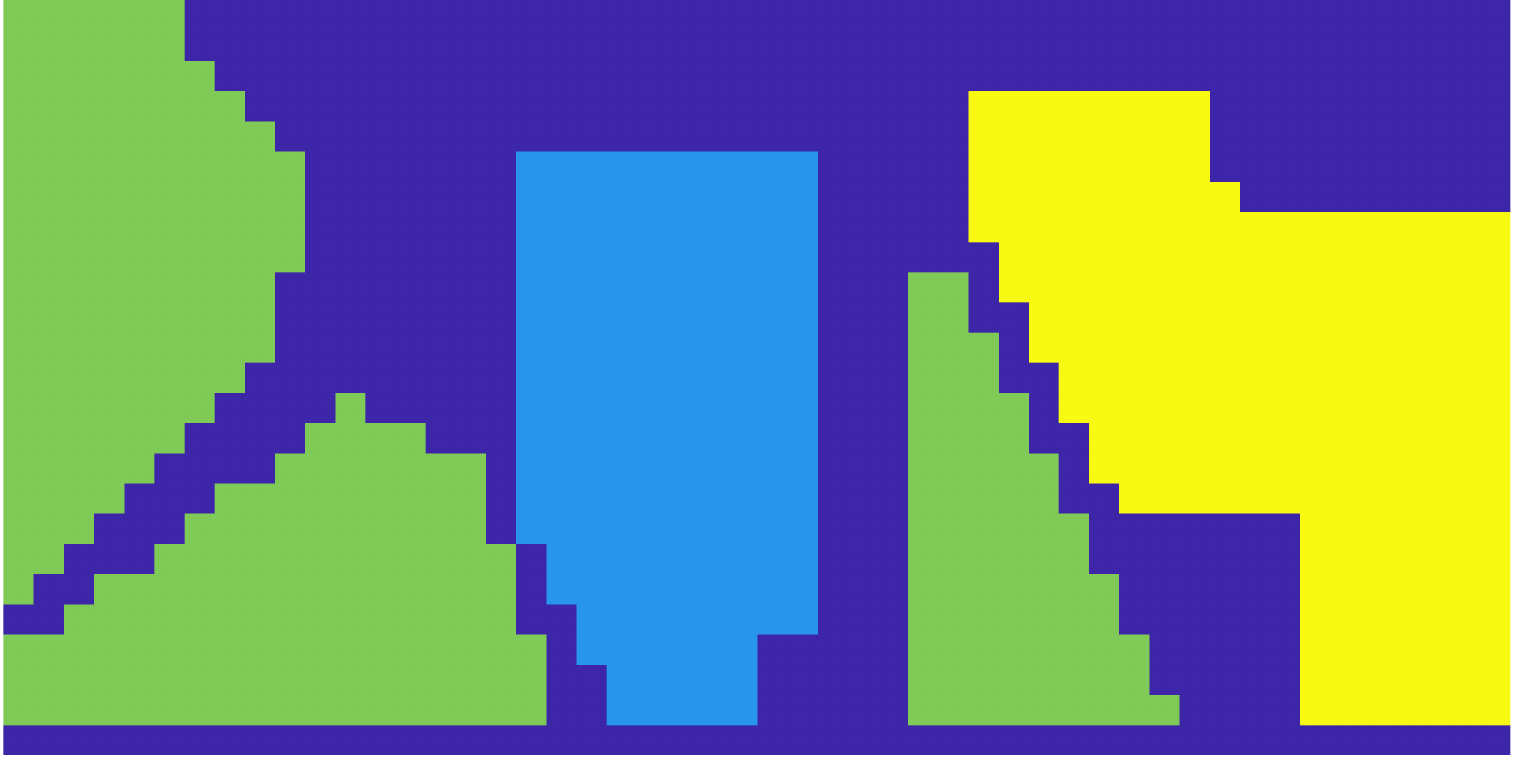}
\caption{\label{fig:IP}The Indian Pines data is a $50\times 25$ subset of the full Indian Pines dataset.  It contains 3 classes, one of which is not well-localized spatially.  The dataset was captured in 1992 in Northwest IN, USA by the AVRIS sensor and has 20m/pixel spatial resolution.  There are $200$ spectral bands. \emph{Left}: projection of the HSI onto its first principal component.  \emph{Right}: ground truth (GT).} 
\end{figure}

\begin{figure}[!htb]
\centering
\begin{subfigure}{ .09\textwidth}
\includegraphics[width=\textwidth]{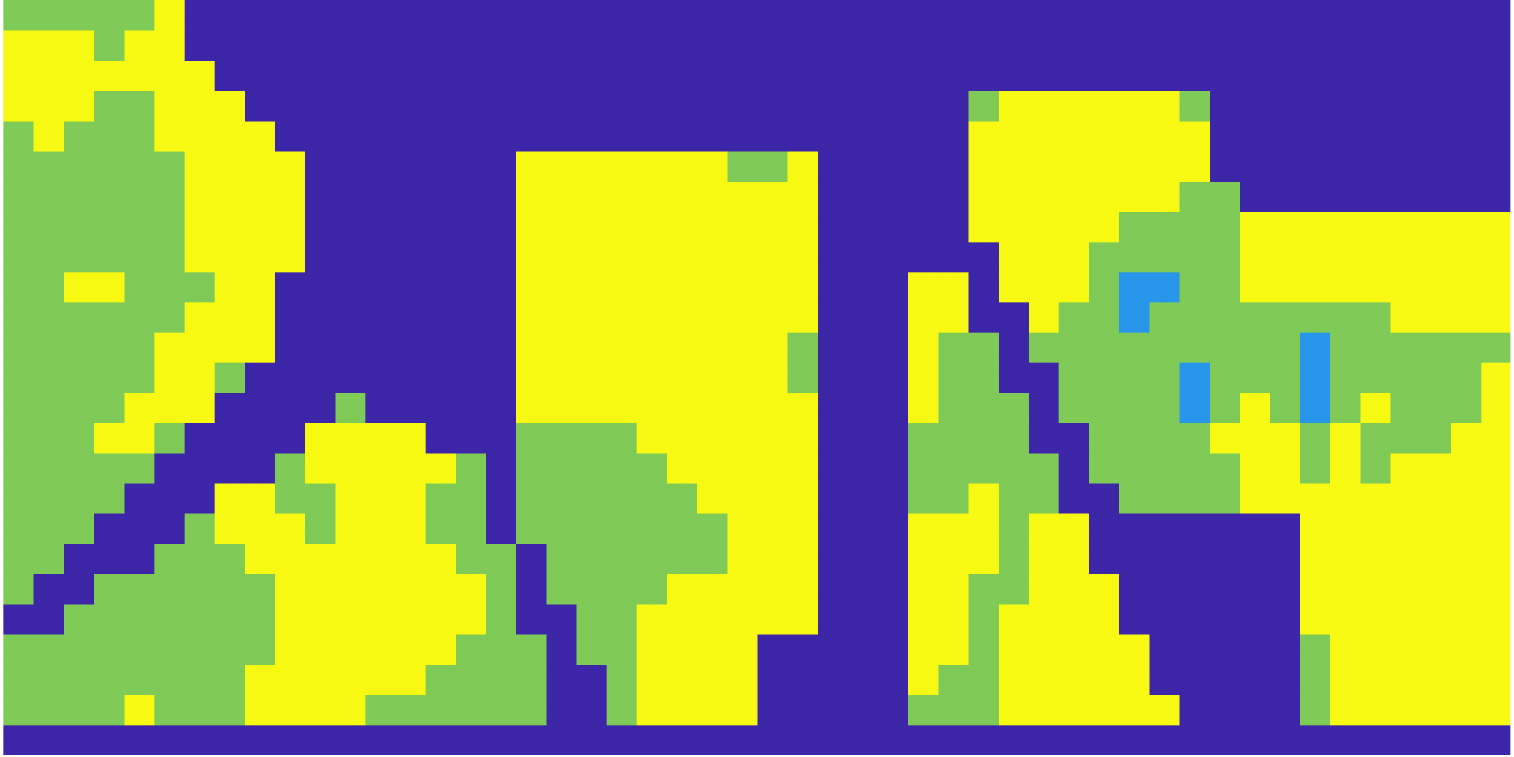}
\caption{$K$-means}
\end{subfigure}
\begin{subfigure}{ .09\textwidth}
\includegraphics[width=\textwidth]{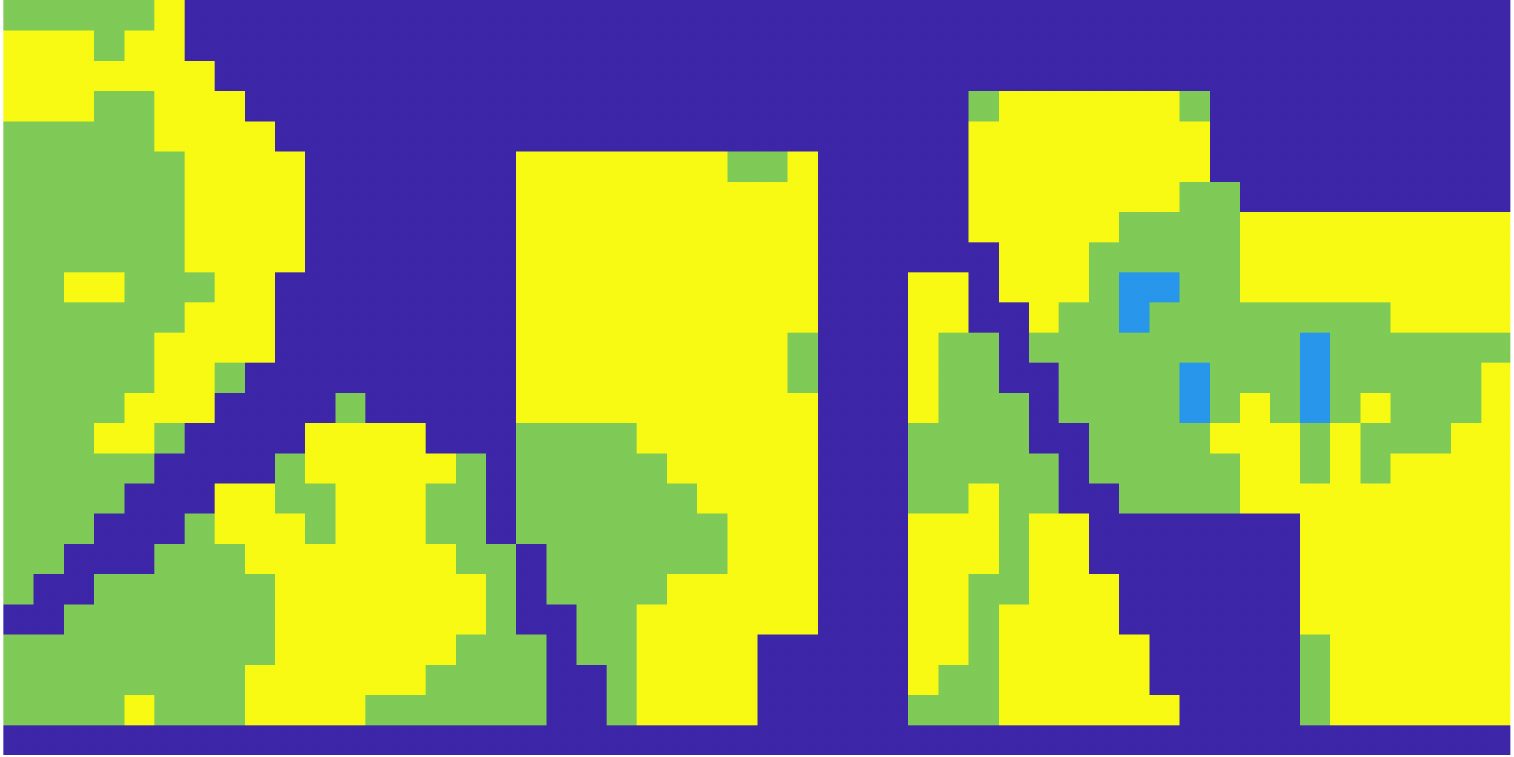}
\caption{PCA+$K$M}
\end{subfigure}
\begin{subfigure}{ .09\textwidth}
\includegraphics[width=\textwidth]{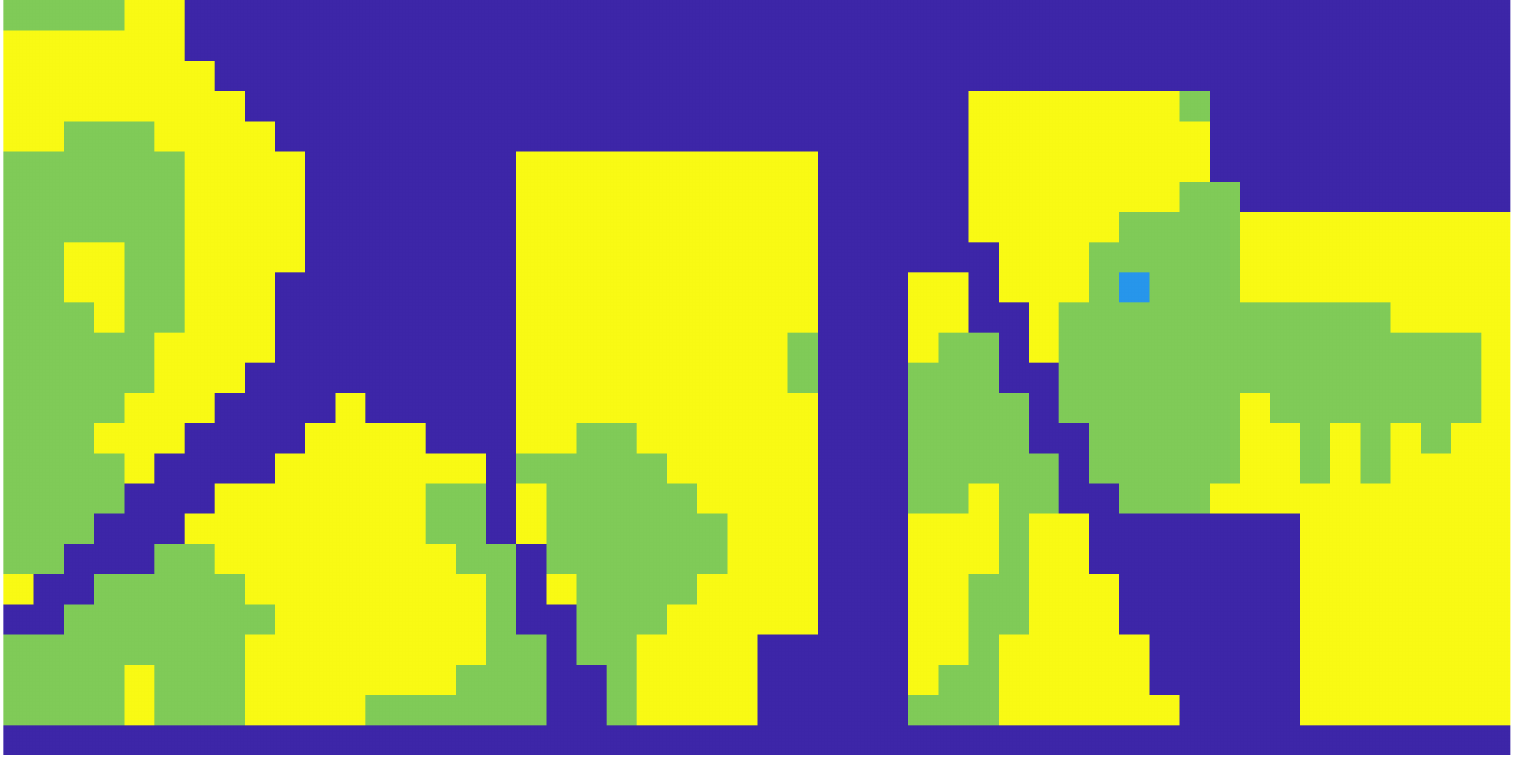}
\caption{ICA+$K$M}
\end{subfigure}
\begin{subfigure}{ .09\textwidth}
\includegraphics[width=\textwidth]{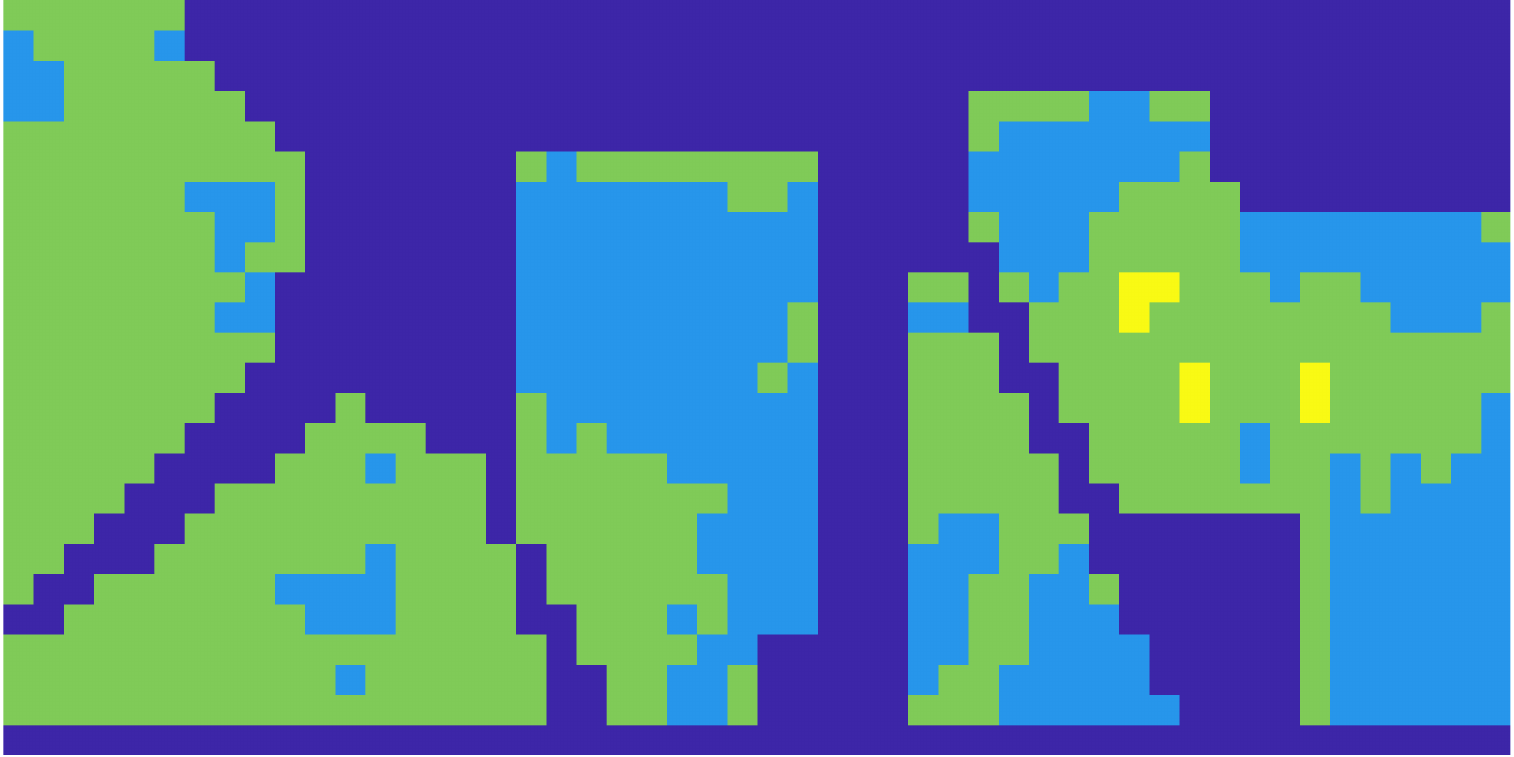}
\caption{RP+$K$M}
\end{subfigure}
\begin{subfigure}{ .09\textwidth}
\includegraphics[width=\textwidth]{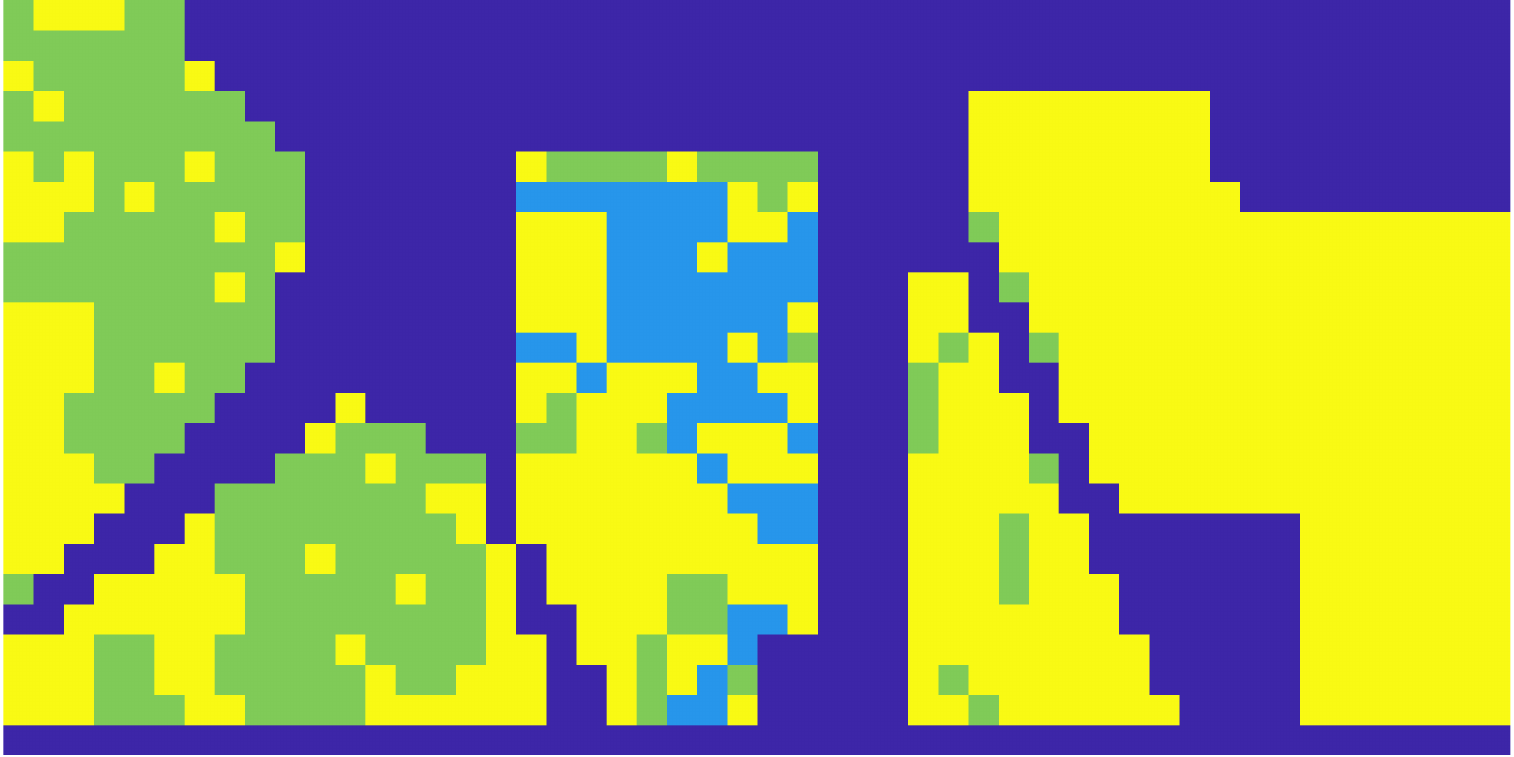}
\caption{DBSCAN}
\end{subfigure}
\begin{subfigure}{ .09\textwidth}
\includegraphics[width=\textwidth]{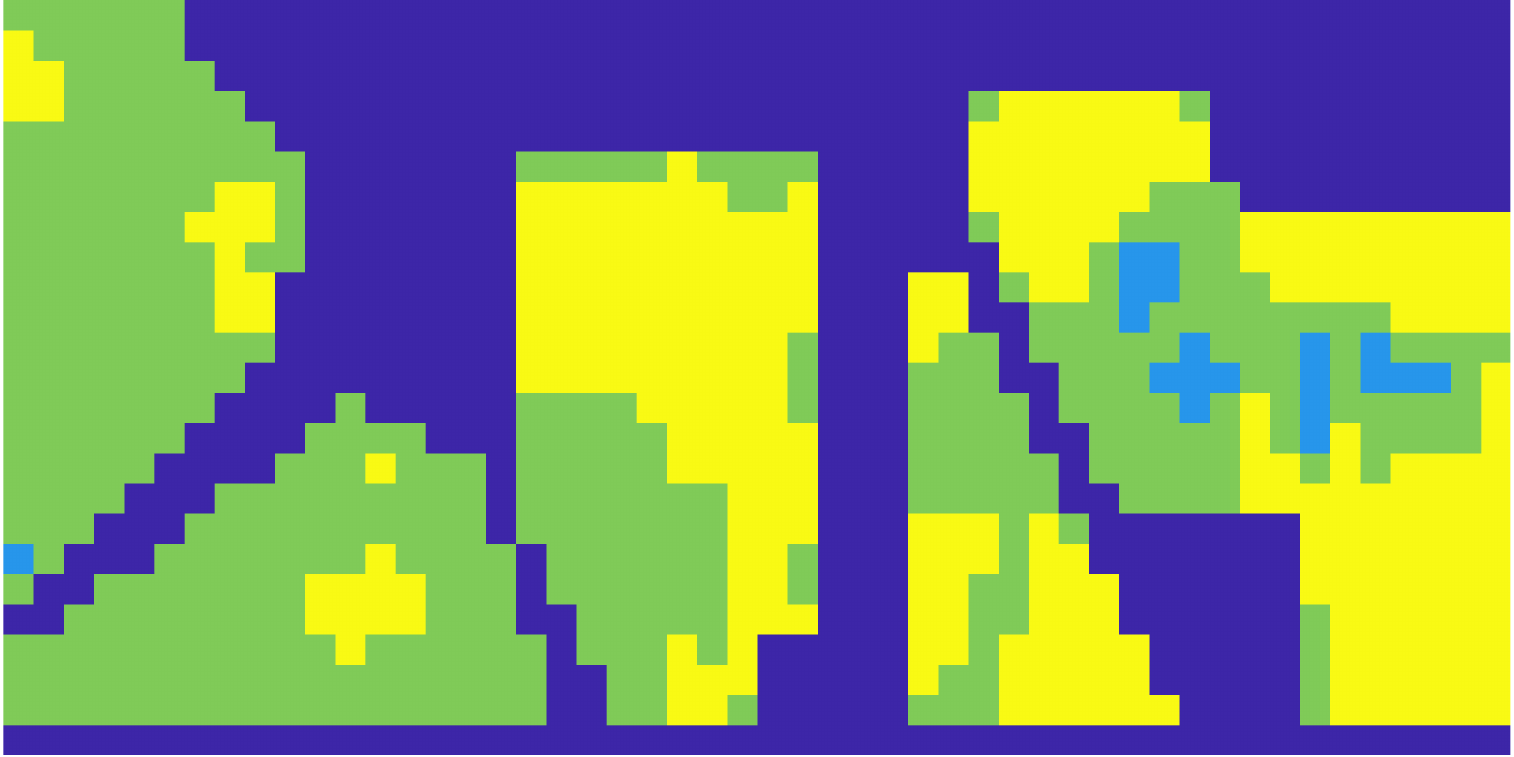}
\caption{SC}
\end{subfigure}
\begin{subfigure}{.09\textwidth}
\includegraphics[width=\textwidth]{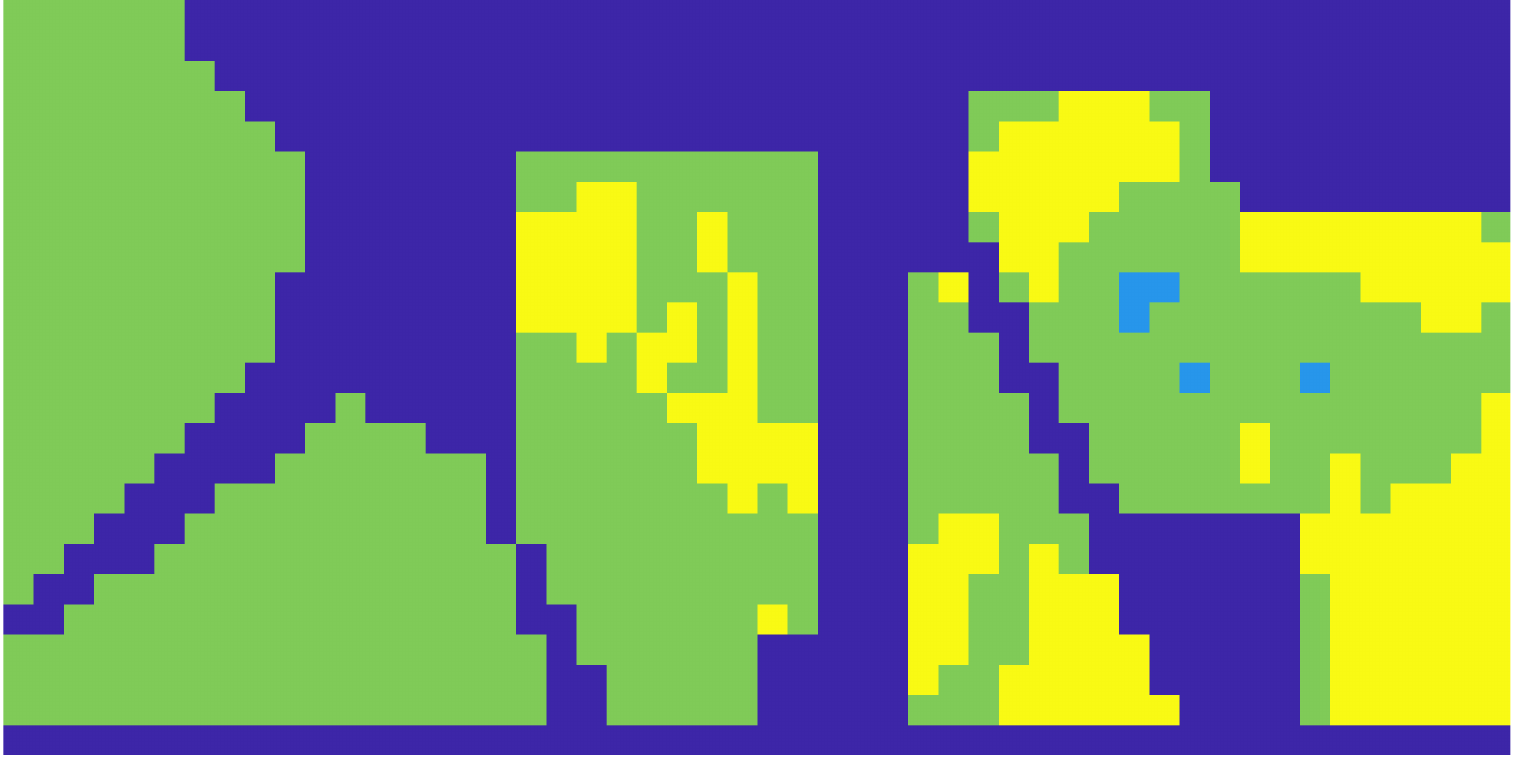}
\caption{GMM}
\end{subfigure}
\begin{subfigure}{.09\textwidth}
\includegraphics[width=\textwidth]{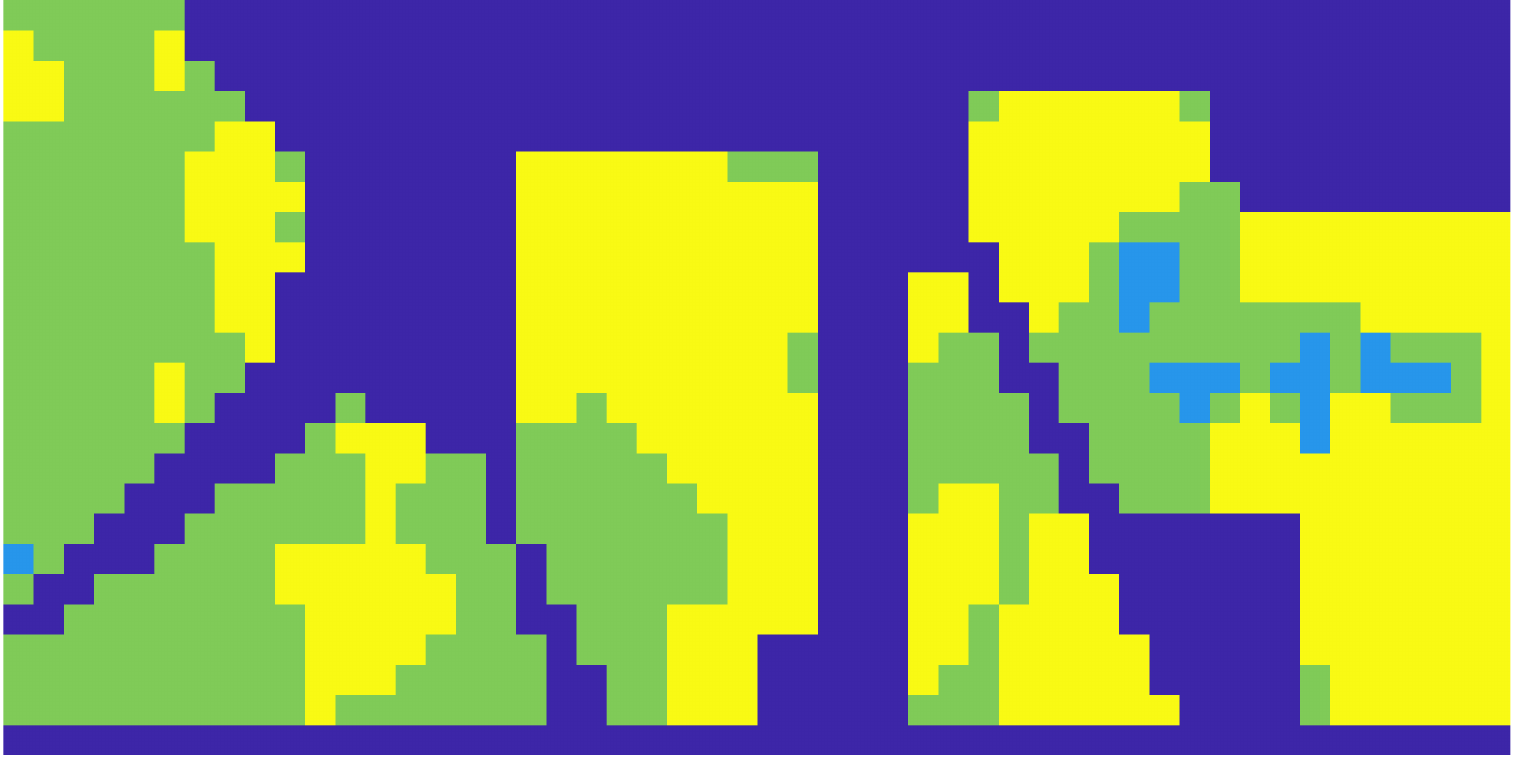}
\caption{SMCE}
\end{subfigure}
\begin{subfigure}{.09\textwidth}
\includegraphics[width=\textwidth]{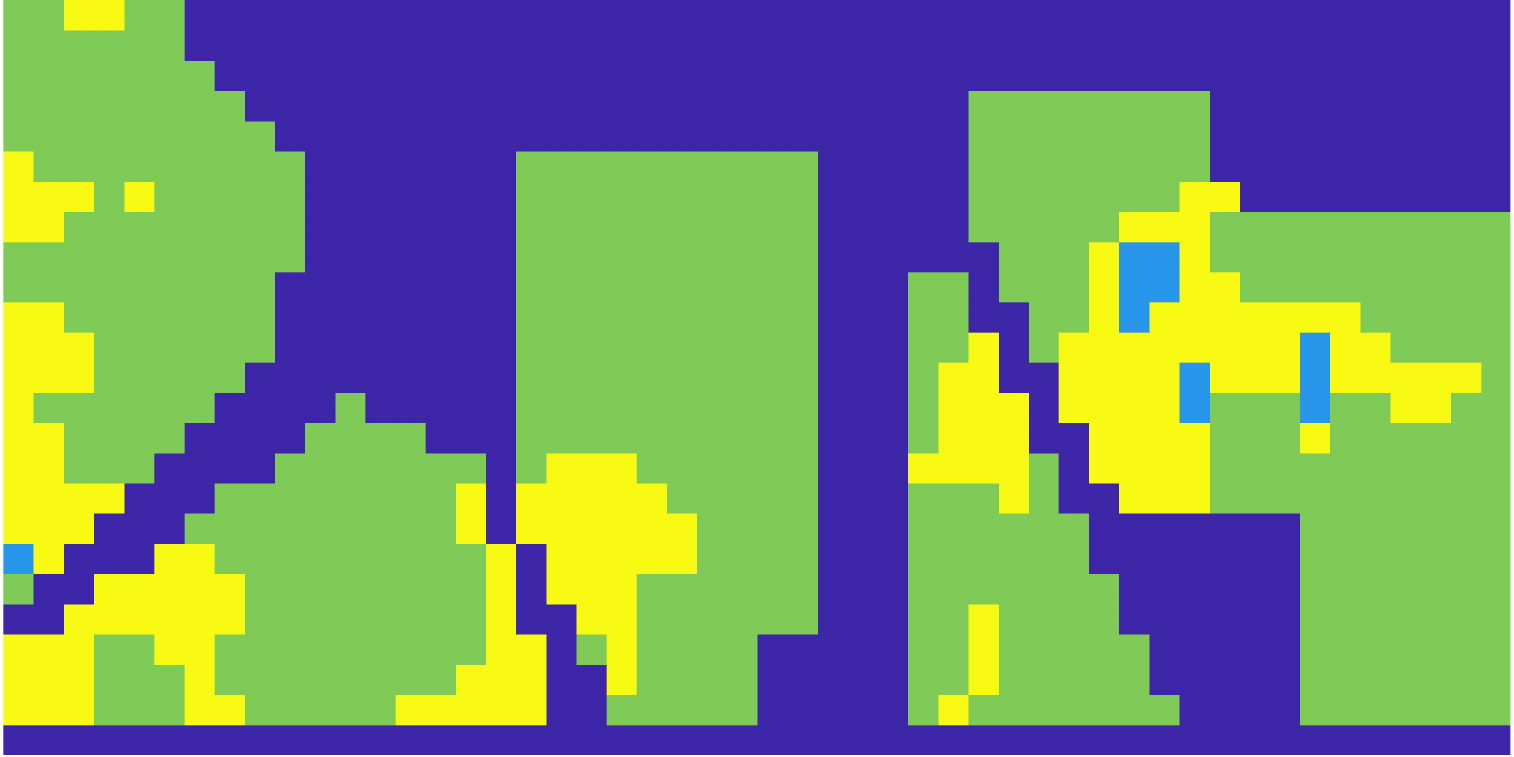}
\caption{HNMF}
\end{subfigure}
\begin{subfigure}{.09\textwidth}
\includegraphics[width=\textwidth]{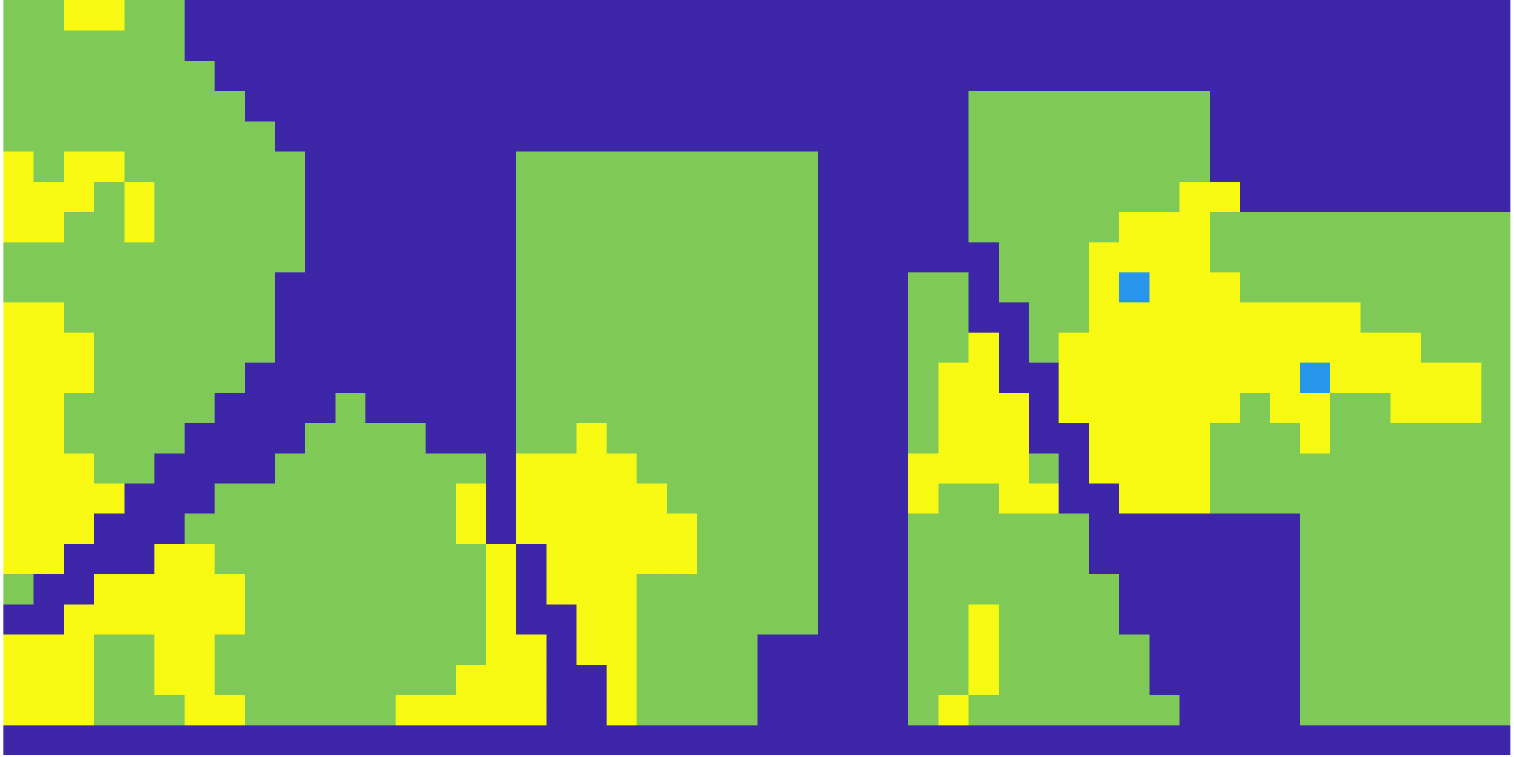}
\caption{FMS}
\end{subfigure}
\begin{subfigure}{ .09\textwidth}
\includegraphics[width=\textwidth]{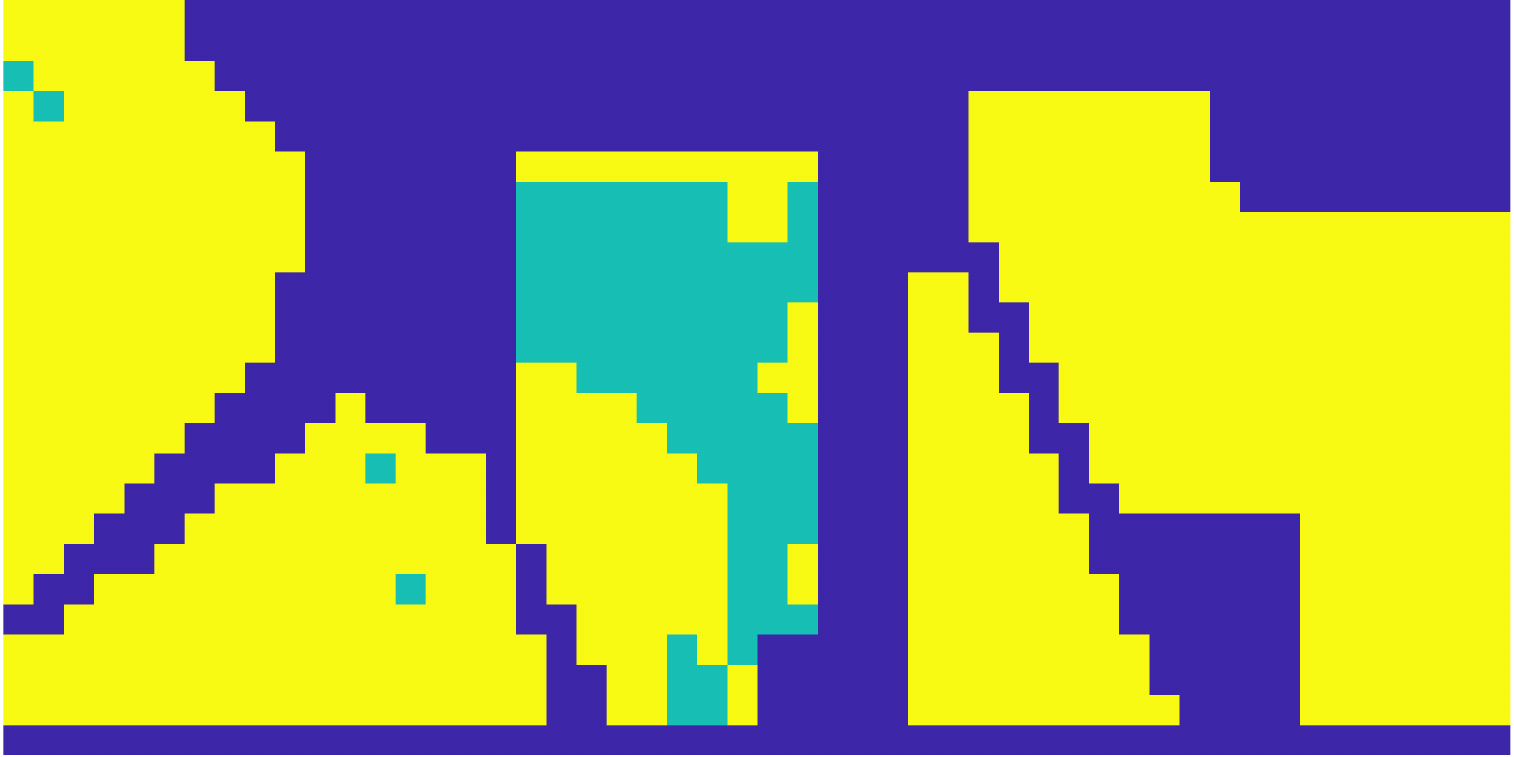}
\caption{FSFDPC}
\end{subfigure}
\begin{subfigure}{ .09\textwidth}
\includegraphics[width=\textwidth]{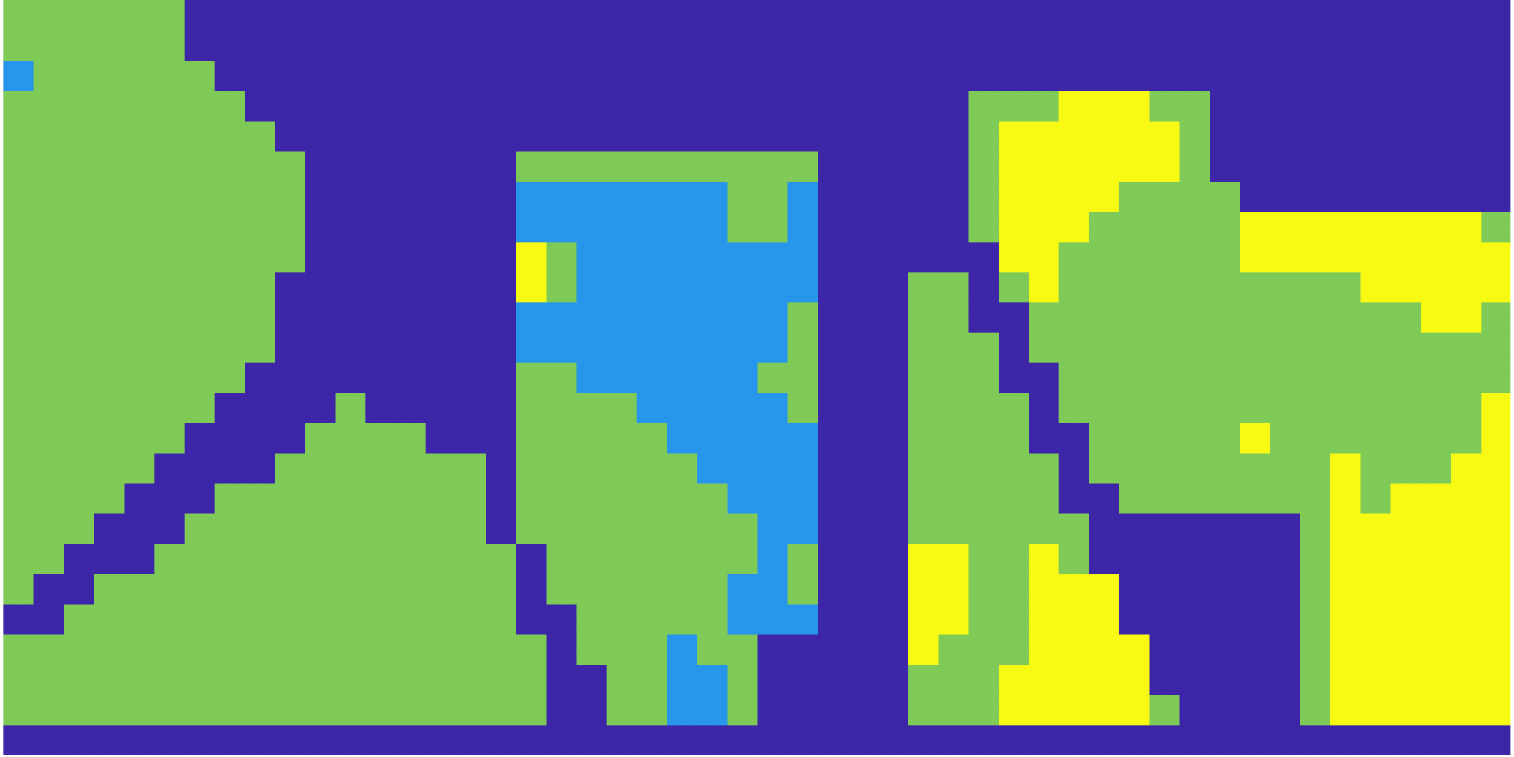}
\caption{DL}
\end{subfigure}
\begin{subfigure}{ .09\textwidth}
\includegraphics[width=\textwidth]{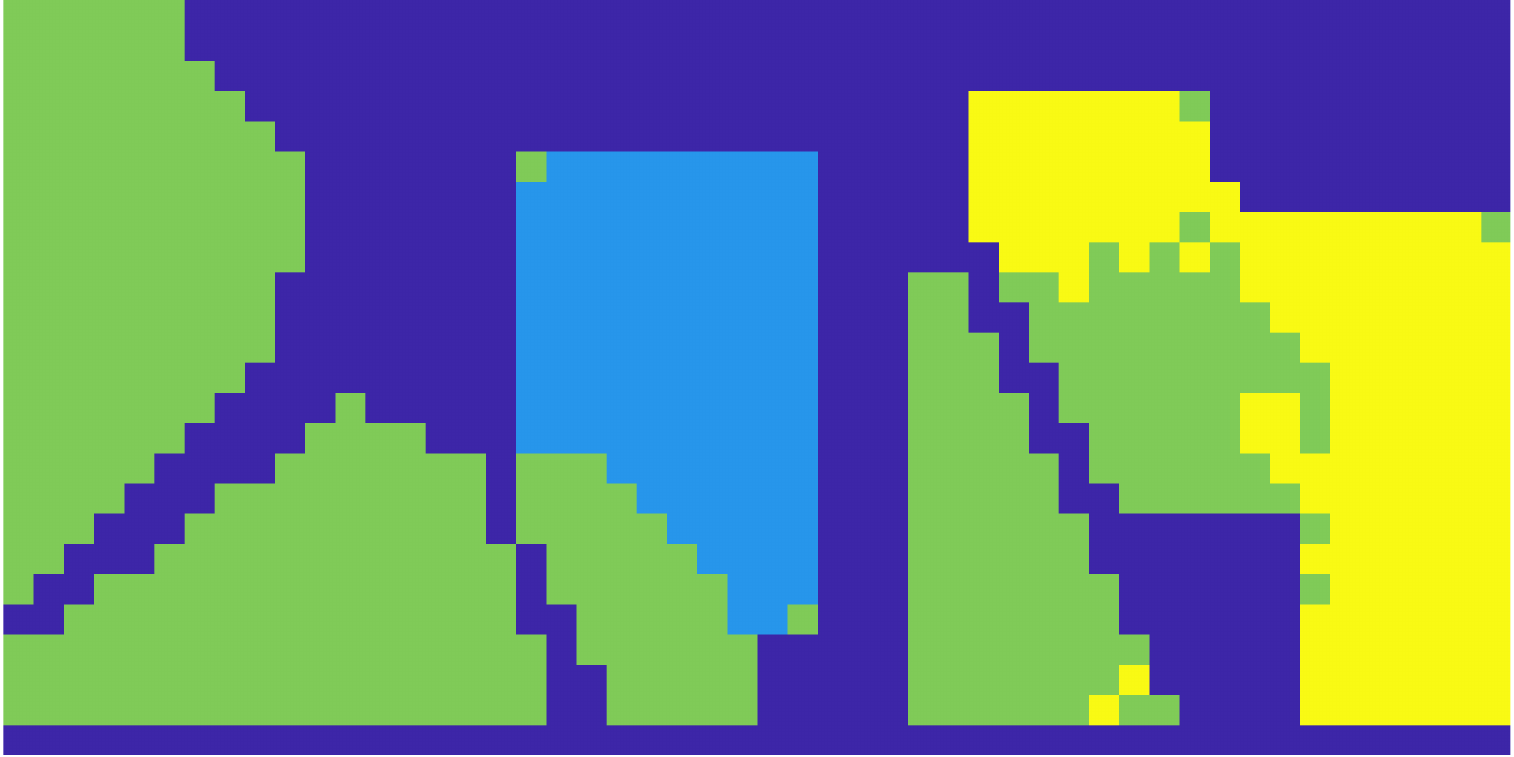}
\caption{DLSS}
\end{subfigure}
\begin{subfigure}{ .09\textwidth}
\includegraphics[width=\textwidth]{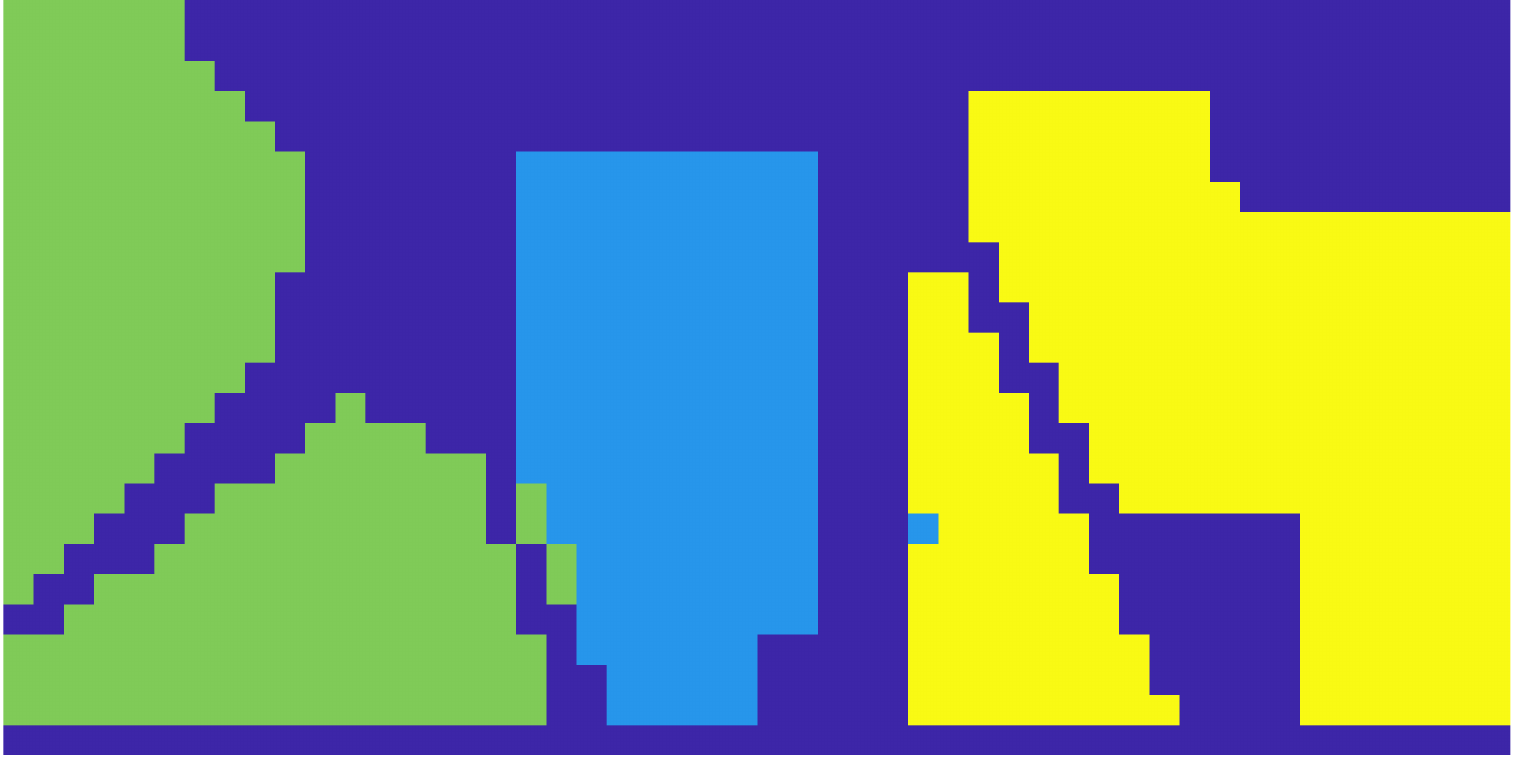}
\caption{SRDL}
\end{subfigure}
\begin{subfigure}{ .09\textwidth}
\includegraphics[width=\textwidth]{Images/IP/IP_GT-crop.pdf}
\caption{GT}
\end{subfigure}
\caption{\label{fig:ResultsIP}Clustering results for Indian Pines dataset.  The SRDL method leads to quite smooth spatial labels, whose accuracy is optimal among all methods.  However, in this case, the ground truth indicates that the triangular region on the lower right is labeled incorrectly by the proposed method.  The smoothing imposed by SRDL---though beneficial overall---washes that region out.  This weakness could be resolved in a variety of ways, most easily perhaps by oversegmenting the HSI, then querying the oversegmented class modes to determine which classes ought to be merged a posteriori.} 
\end{figure}

The Indian Pines dataset used for experiments is a subset of the full Indian Pines datasets, consisting of three classes that are difficult to distinguish visually; see Figure \ref{fig:IP}.   Clustering results for Indian Pines are in Figure \ref{fig:ResultsIP} and Table \ref{tab:Summary}.  SRDL was run with $r=8$.  The spatial regularization in the construction of $P$ is beneficial in terms of labeling: the proposed method improves over DLSS.  As expected, SRDL leads to a spatially smoother labeling.  However, a mistake is still made in the labeling of the proposed method, indicating that this is a challenging dataset to cluster without supervision.  

\subsection{Salinas A Data}

\begin{figure}[!htb]
\centering
\begin{subfigure}{.09\textwidth}
\includegraphics[width=\textwidth]{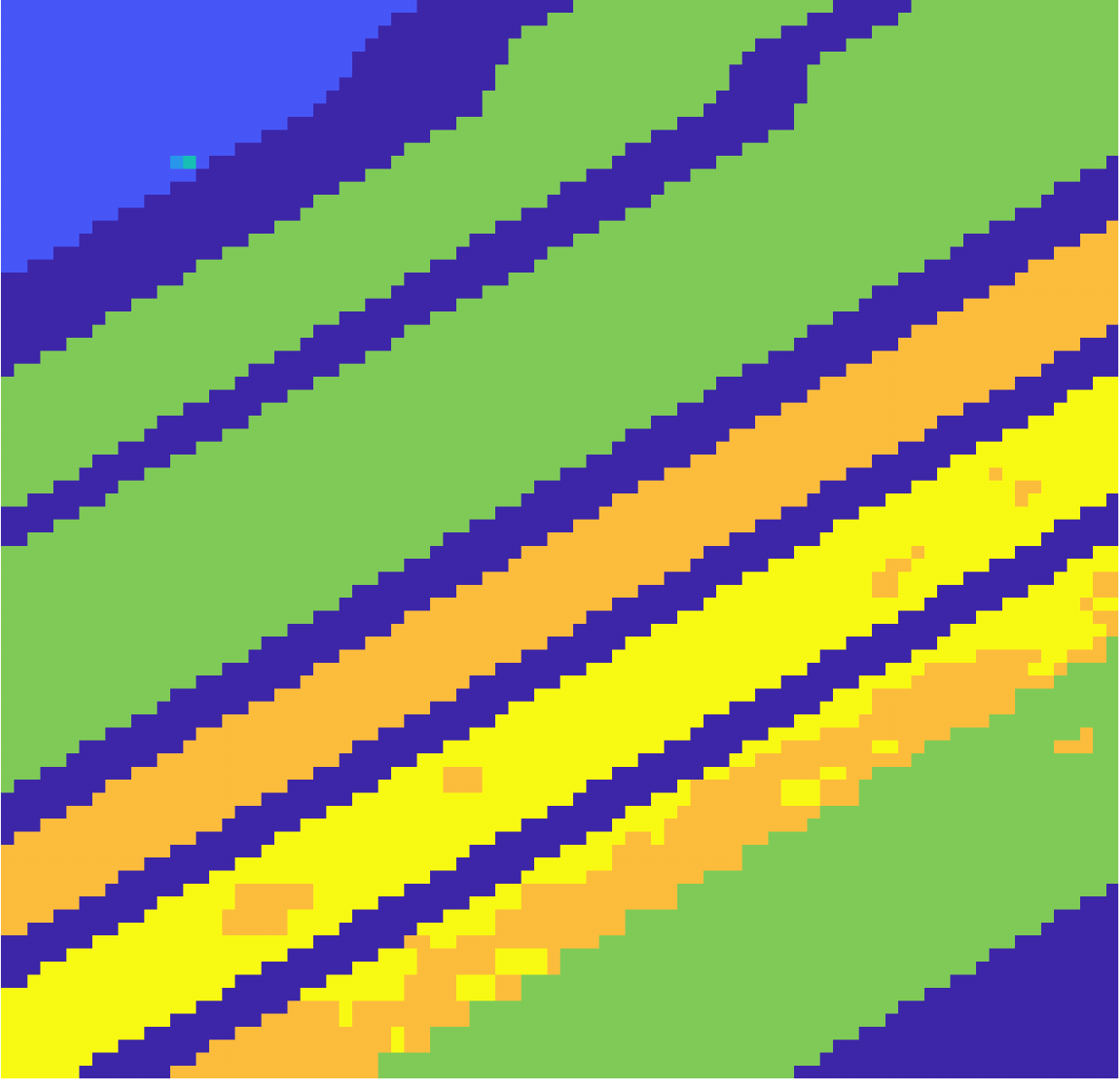}
\caption{$K$-means}
\end{subfigure}
\begin{subfigure}{.09\textwidth}
\includegraphics[width=\textwidth]{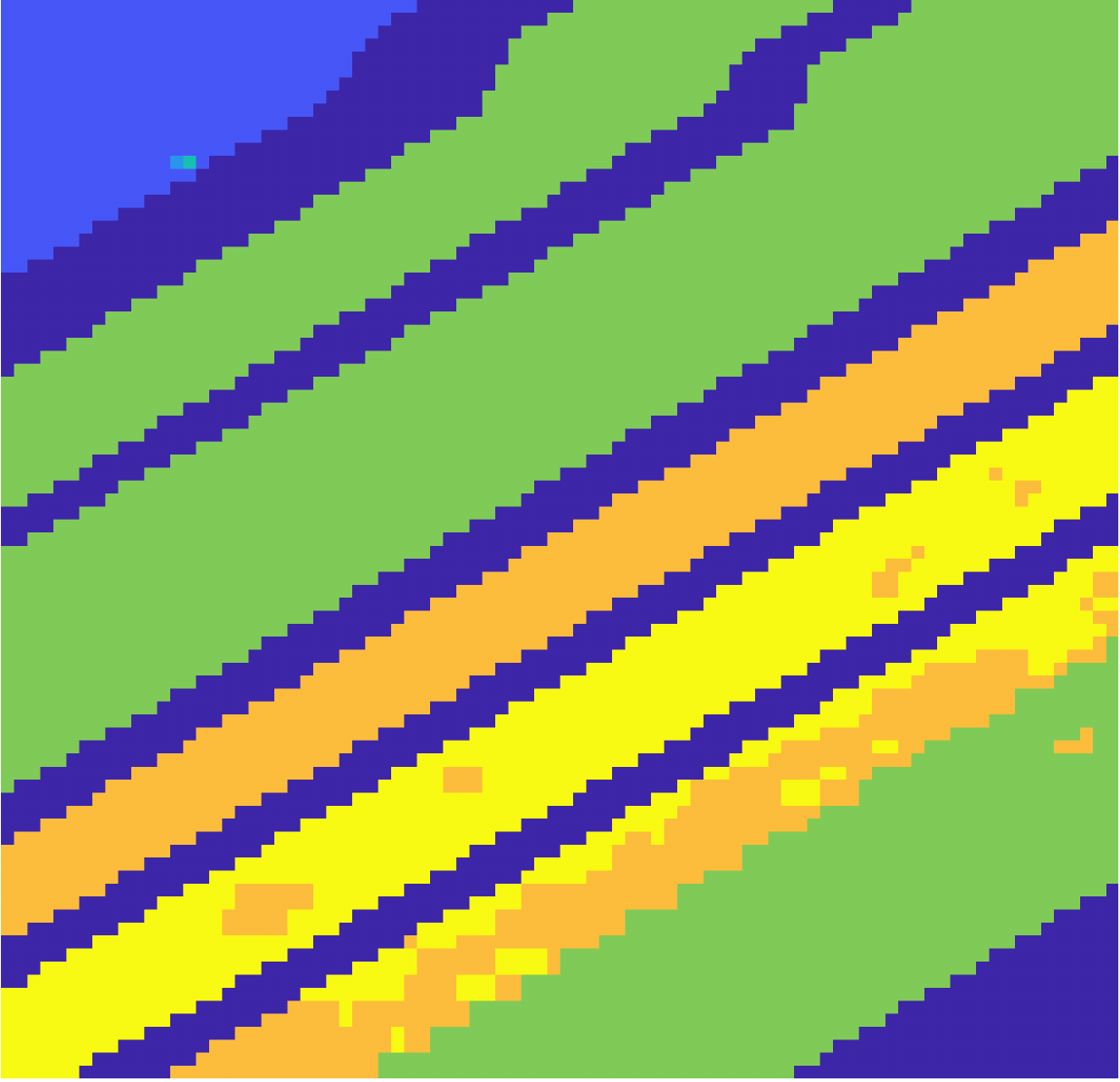}
\caption{PCA+$K$M}
\end{subfigure}
\begin{subfigure}{.09\textwidth}
\includegraphics[width=\textwidth]{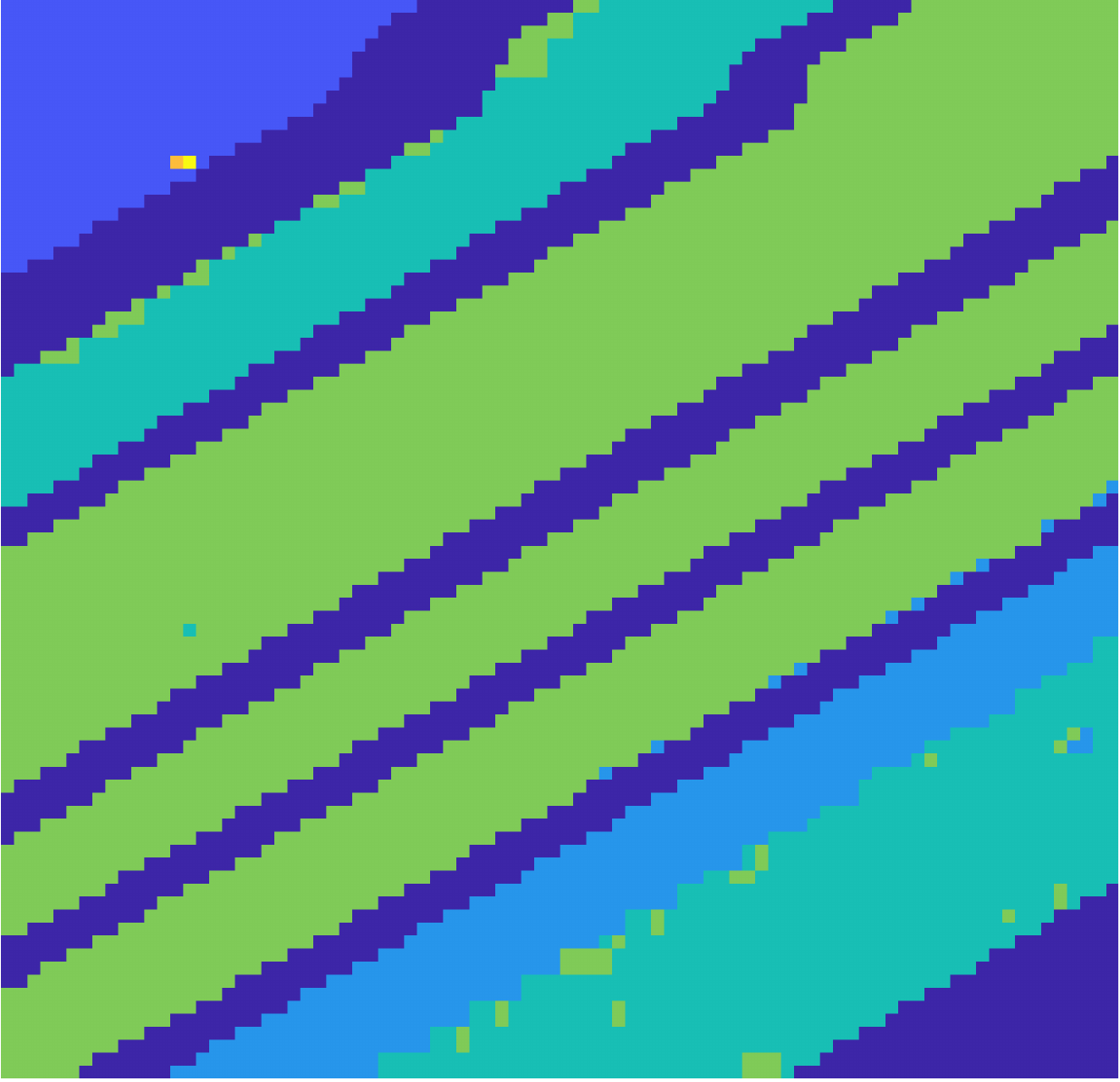}
\caption{ICA+$K$M}
\end{subfigure}
\begin{subfigure}{.09\textwidth}
\includegraphics[width=\textwidth]{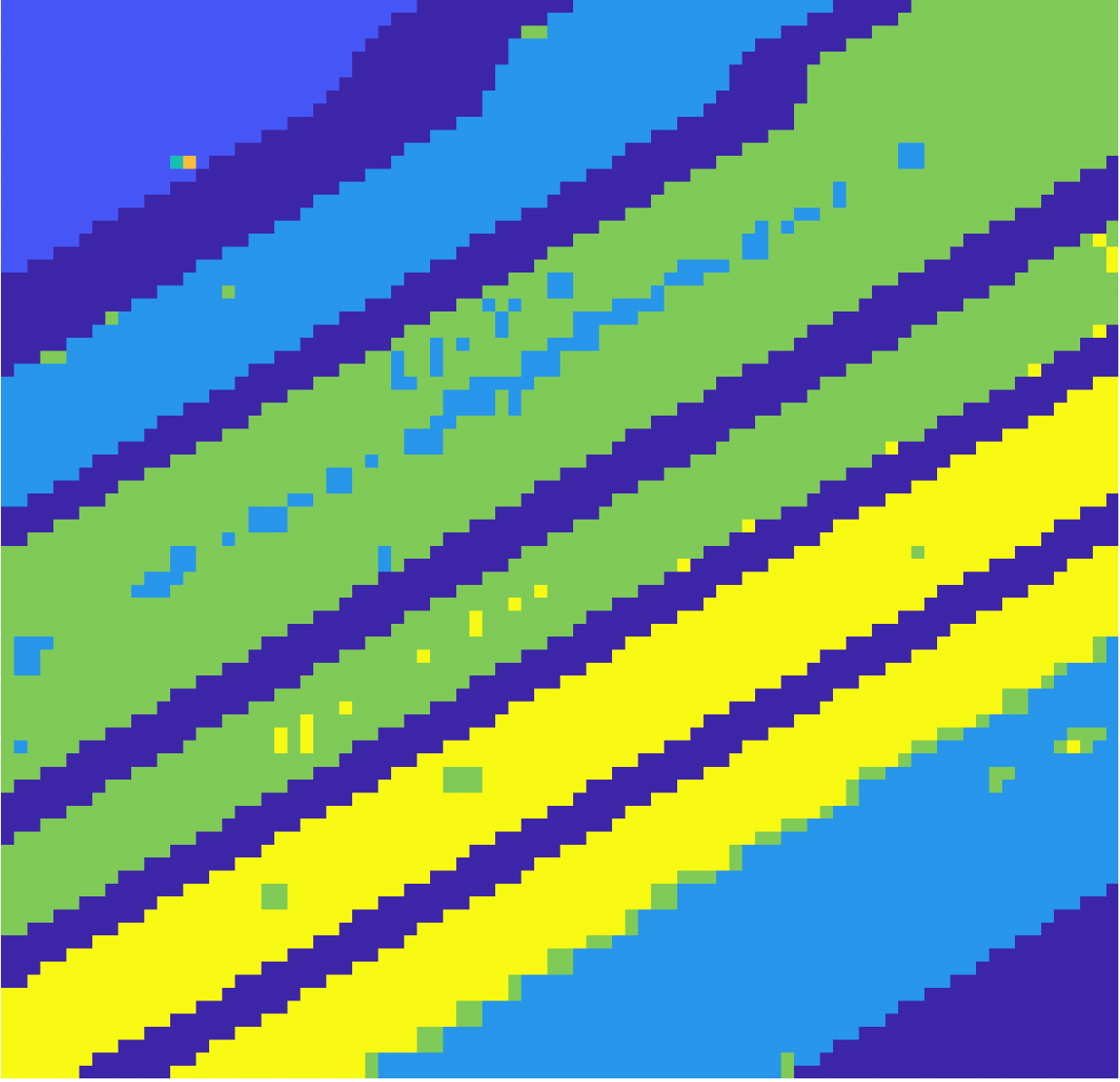}
\caption{RP+$K$M}
\end{subfigure}
\begin{subfigure}{ .09\textwidth}
\includegraphics[width=\textwidth]{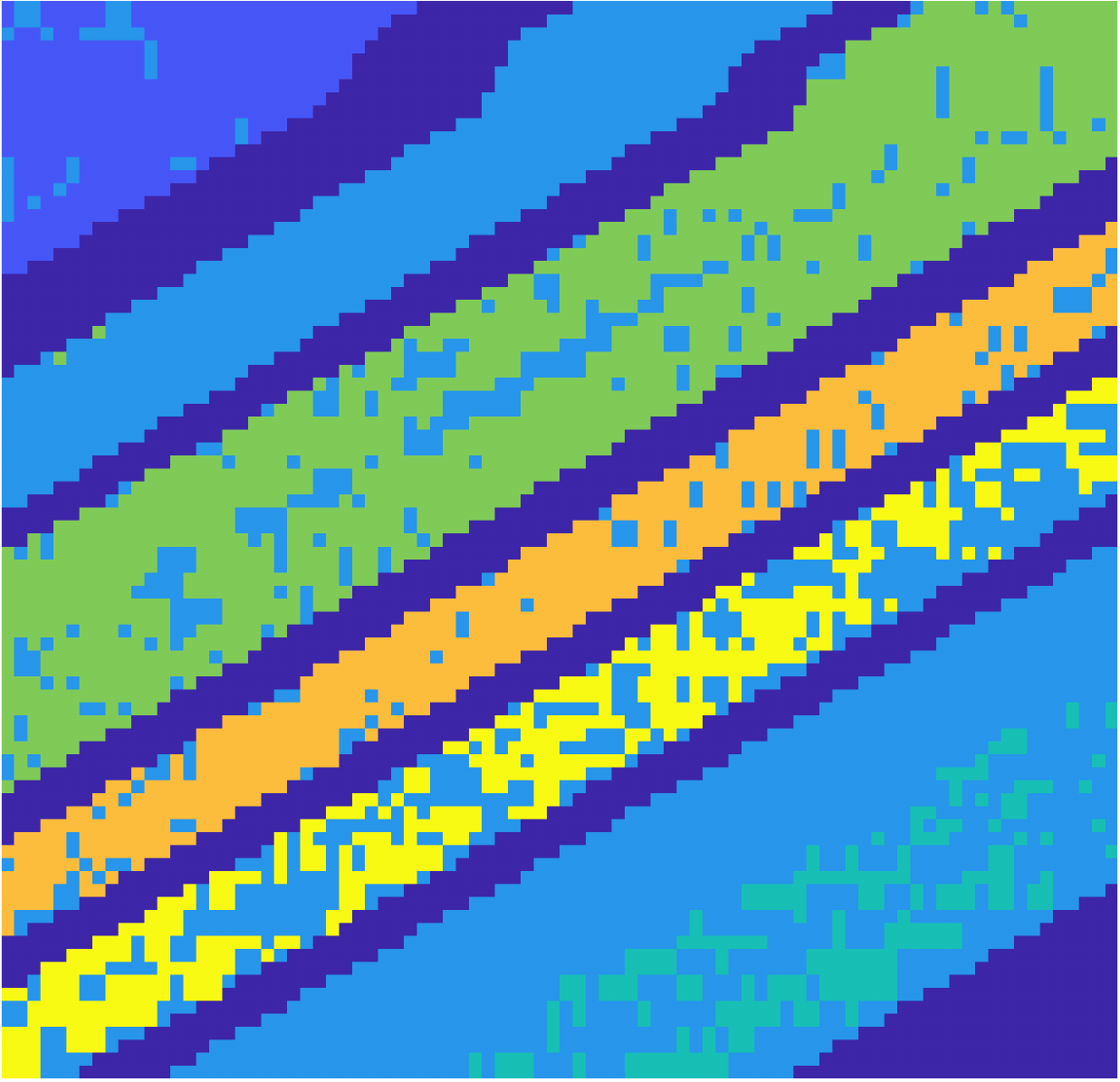}
\caption{DBSCAN}
\end{subfigure}
\begin{subfigure}{.09\textwidth}
\includegraphics[width=\textwidth]{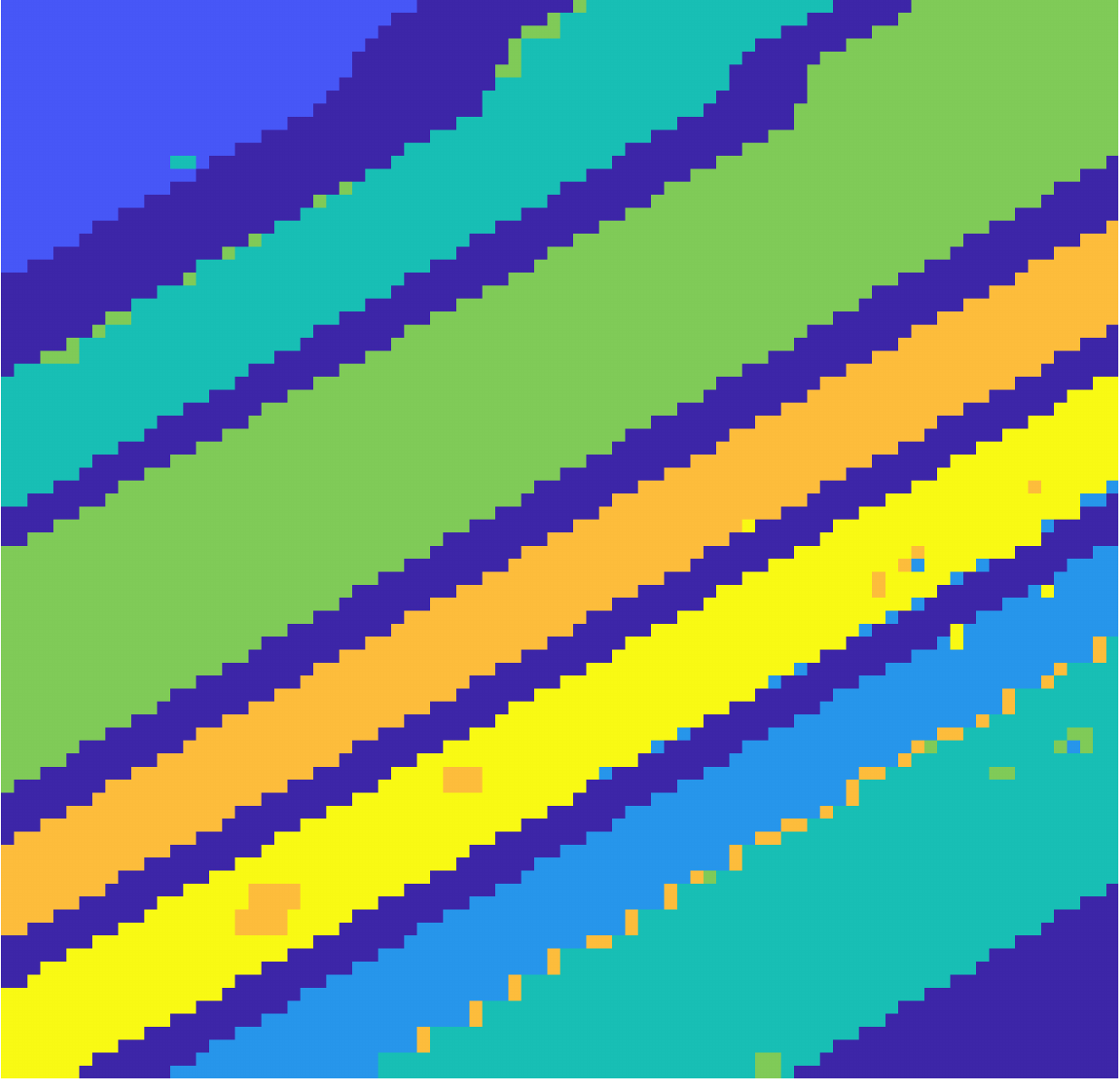}
\caption{SC}
\end{subfigure}
\begin{subfigure}{.09\textwidth}
\includegraphics[width=\textwidth]{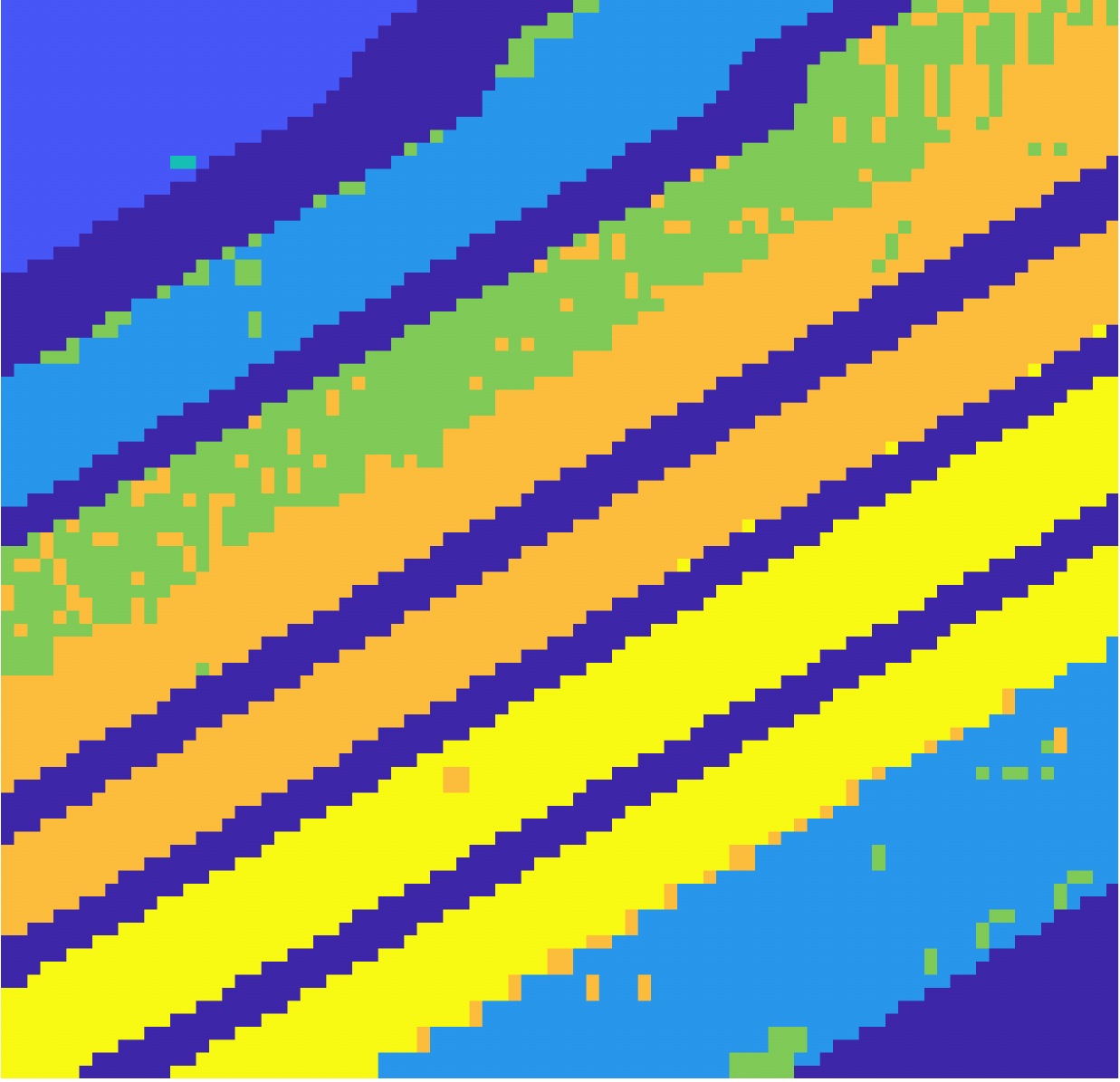}
\caption{GMM}
\end{subfigure}
\begin{subfigure}{.09\textwidth}
\includegraphics[width=\textwidth]{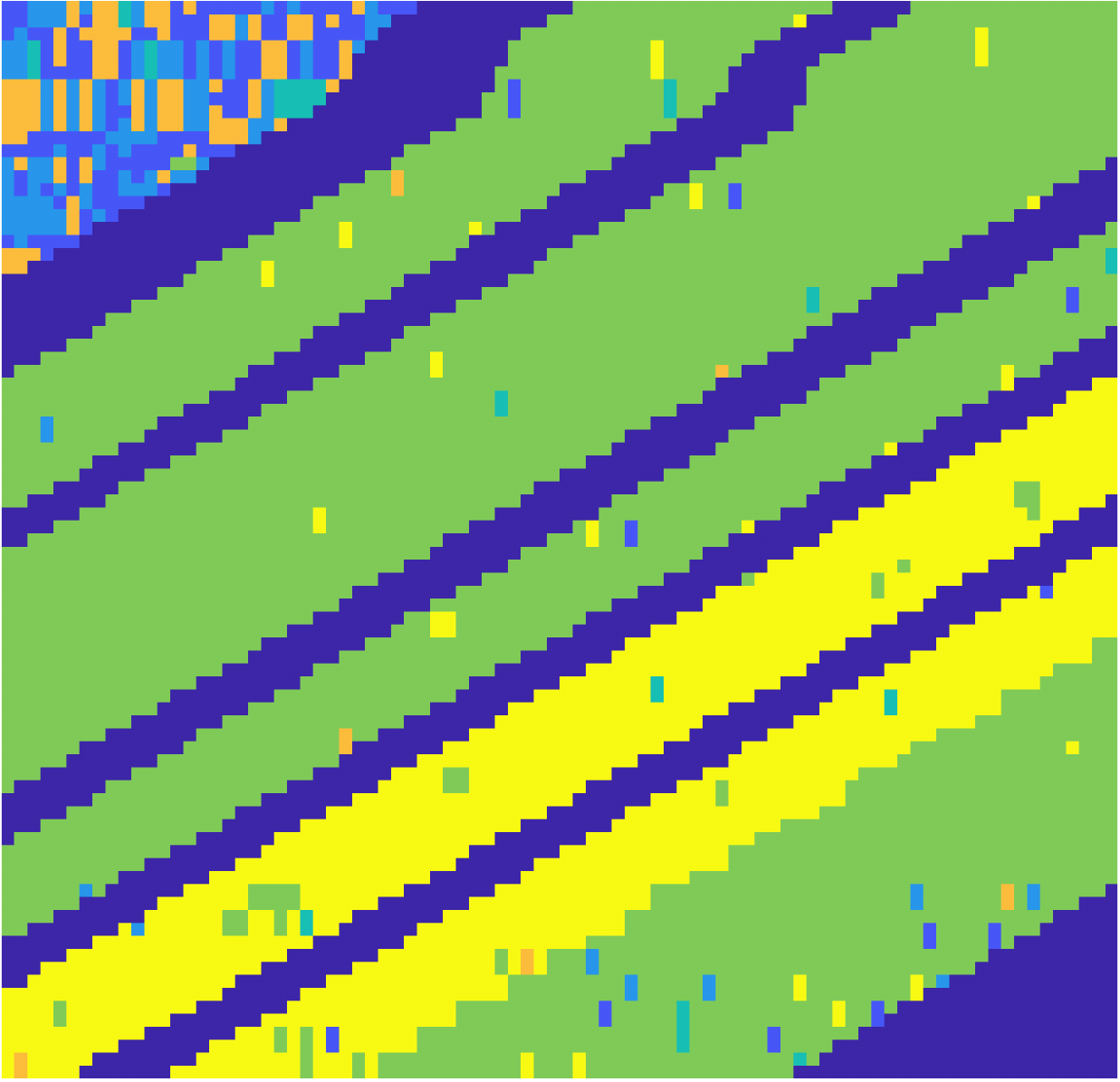}
\caption{SMCE}
\end{subfigure}
\begin{subfigure}{.09\textwidth}
\includegraphics[width=\textwidth]{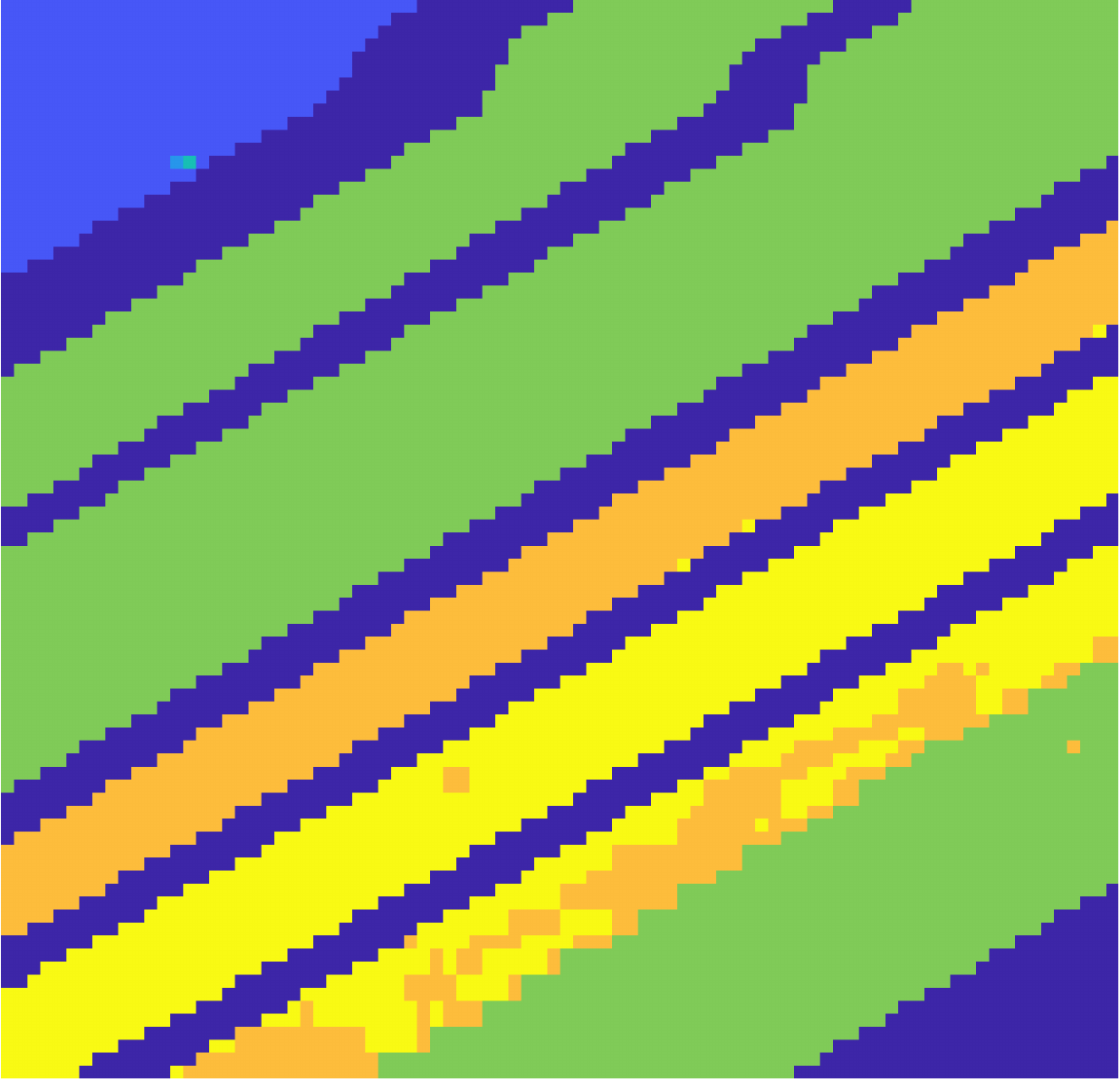}
\caption{HNMF}
\end{subfigure}
\begin{subfigure}{ .09\textwidth}
\includegraphics[width=\textwidth]{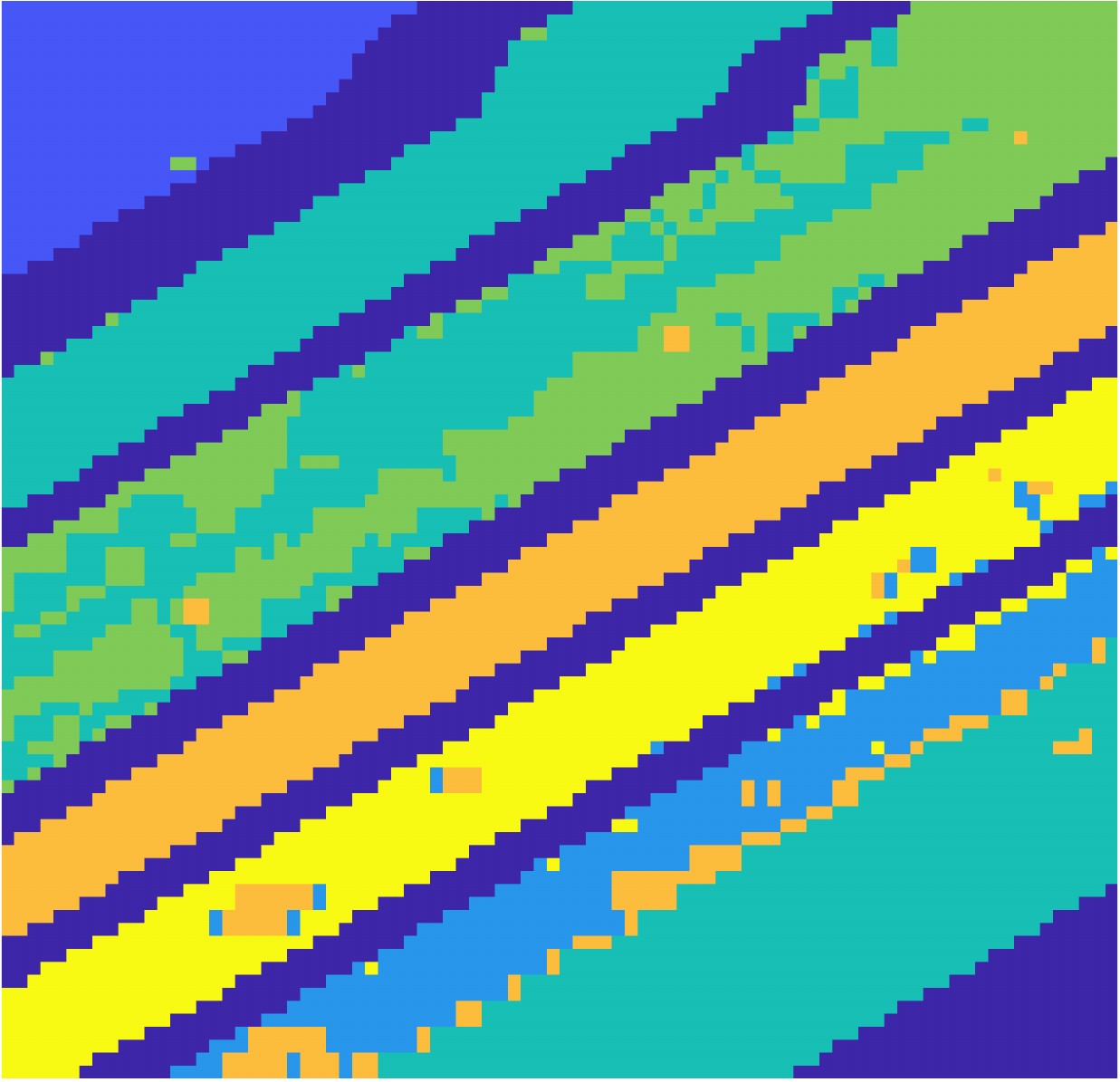}
\caption{FMS}
\end{subfigure}
\begin{subfigure}{ .09\textwidth}
\includegraphics[width=\textwidth]{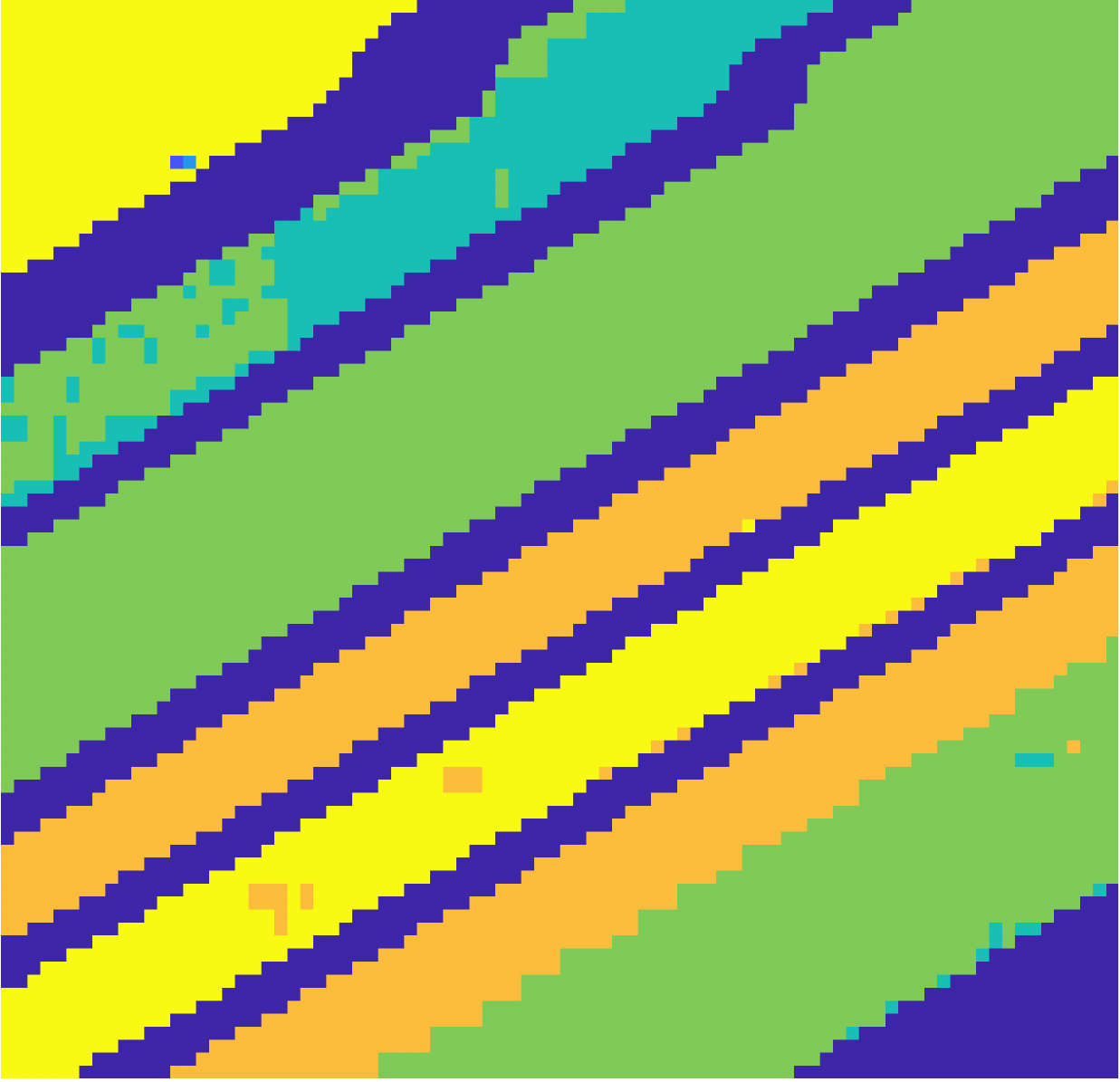}
\caption{FSFDPC}
\end{subfigure}
\begin{subfigure}{ .09\textwidth}
\includegraphics[width=\textwidth]{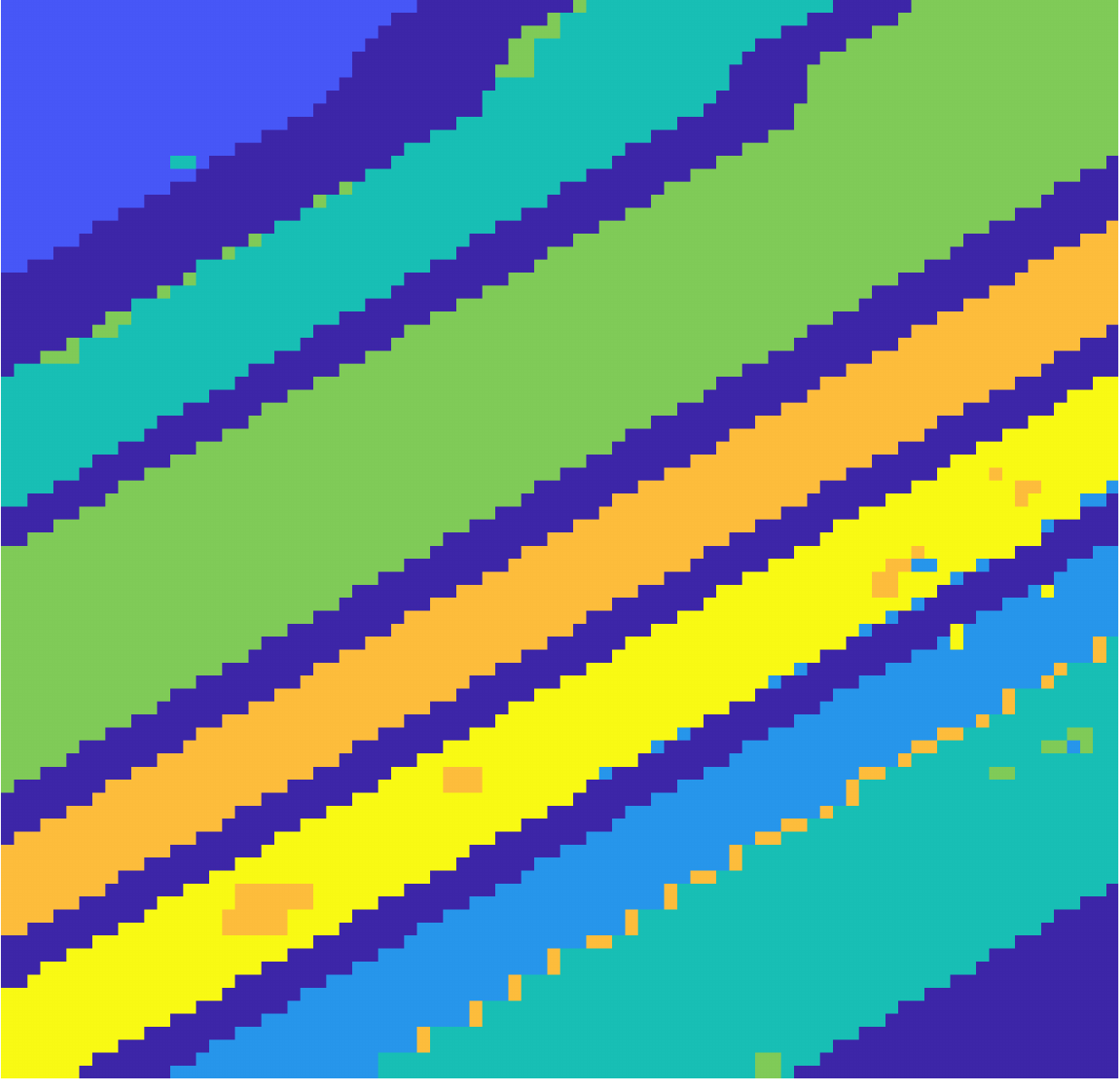}
\caption{DL}
\end{subfigure}
\begin{subfigure}{ .09\textwidth}
\includegraphics[width=\textwidth]{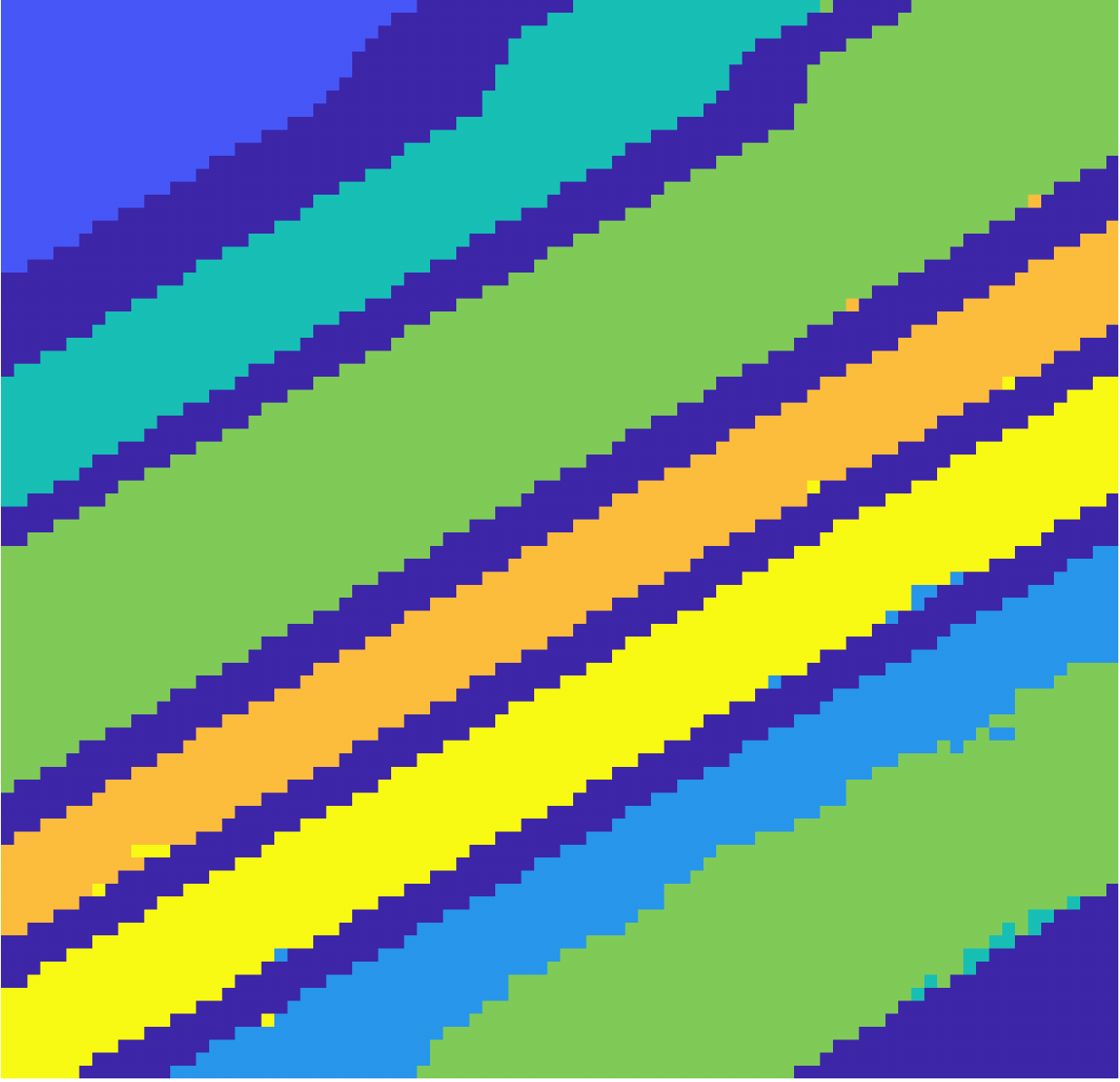}
\caption{DLSS}
\end{subfigure}
\begin{subfigure}{ .09\textwidth}
\includegraphics[width=\textwidth]{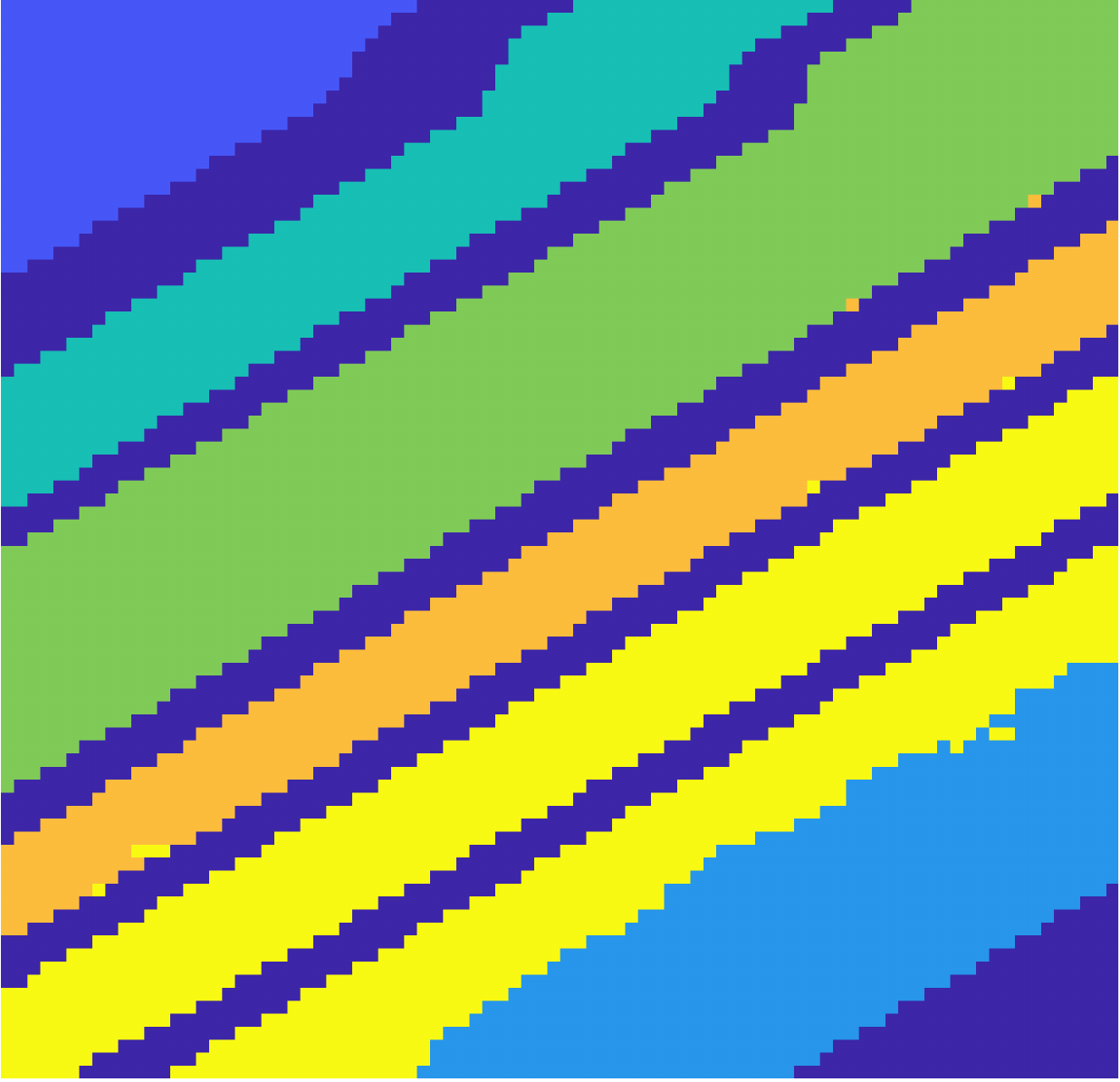}
\caption{SRDL}
\end{subfigure}
\begin{subfigure}{ .09\textwidth}
\includegraphics[width=\textwidth]{Images/SalinasA/SalinasA_GT-crop.pdf}
\caption{GT}
\end{subfigure}
\caption{\label{fig:ResultsSalinasA}Clustering results for Salinas A dataset.  The proposed method is the optimal performer, with the DLSS, DL, and spectral clustering methods also performing strongly.  The spatial regularization incorporated into the diffusion distances used for the proposed method keeps the diagonal stripes relatively far apart from each other, leading to accurate mode estimate and good subsequent labeling.} 
\end{figure}

The Salinas A dataset (see Figure \ref{fig:SpatialNeighbors}) consists of 6 classes in diagonal rows.  Certain pixels in the HSI have the same values; in order to distinguish these pixels, small Gaussian noise (variance $< 10^{-3})$ was added as a preprocessing step.   SRDL was run with $r=20$.  Visual results for Salinas A appear in Figure \ref{fig:ResultsSalinasA} and quantitative results appear in Table \ref{tab:Summary}.  The proposed method yields the best results, and moreover the labels recovered by the proposed method are quite spatially regular.  

\subsection{Kennedy Space Center Data}

\begin{figure}[!htb]
\centering
\includegraphics[width=.24\textwidth]{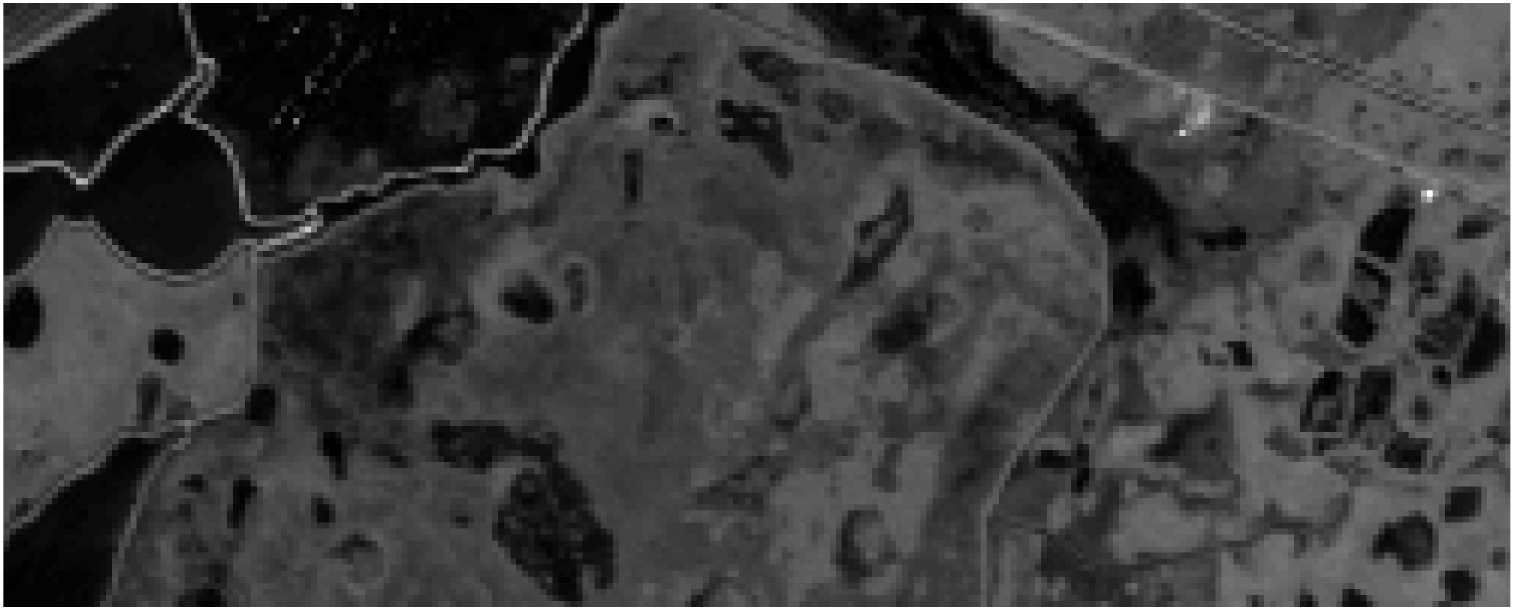}
\includegraphics[width=.24\textwidth]{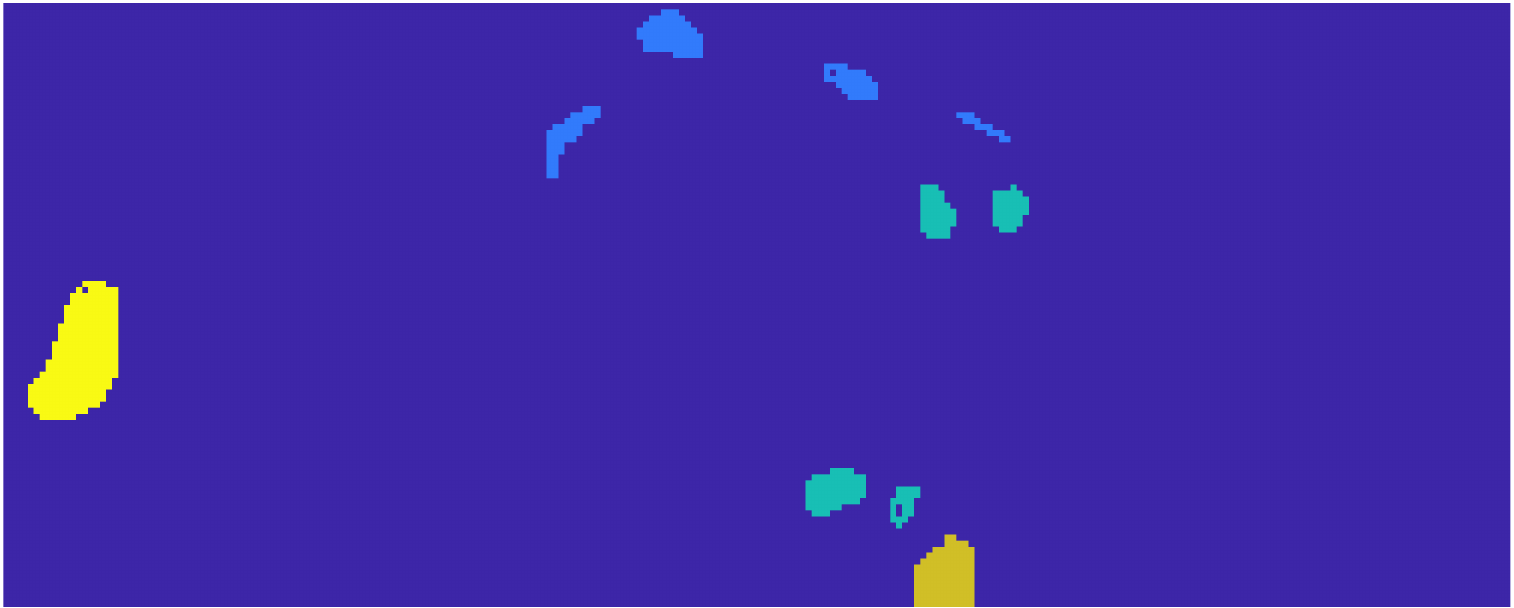}
\caption{\label{fig:KSC}The Kennedy Space Center data is a $250\times 100$ subset of the full Kennedy Space Center dataset.  The scene was captured with the NASA AVIRIS instrument over the Kennedy Space Center (KSC), Florida, USA and has 18m/pixel spatial resolution.  It consists of 4 classes, some of which have poor spatial localization.  The dataset consists of 176 bands after removing low signal-to-noise-ratio and water-absorption bands.  \emph{Left}: projection of the data onto its first principal component.  \emph{Right}: ground truth (GT).} 
\end{figure}

The Kennedy Space Center dataset used for experiments consists of a subset of the original dataset, and contains four classes.  Figure \ref{fig:KSC} shows the data along with ground truth consisting of the examples of four vegetation types which dominate the scene.   SRDL was run with $r=20$.  Results appear in Table \ref{tab:Summary}; for reasons of space, visual results are not shown.  

\begin{table}[htb!]
\begin{adjustbox}{max width=.5\textwidth}
\begin{tabular}{| c | c | c | c | c | c | c  | c | c | c |}\hline
Method & OA I.P. & AA I.P. & $\kappa$ I.P. & OA S.A. & AA S.A. & $\kappa$ S.A. & OA  K.S.C. & AA K.S.C. & $\kappa$ K.S.C.\\ \hline
$K$-means & 0.43 & 0.38 & 0.09 & 0.63 & 0.66 & 0.52 & 0.36 & 0.25 & 0.01 \\ \hline
PCA+$K$-means & 0.43  & 0.38  & 0.10 &  0.63 &  0.66 &  0.52 & 0.36  & 0.25 & 0.01 \\ \hline
ICA+$K$-means & 0.41 & 0.36 & 0.06 &  0.57 &  0.56 & 0.44  & 0.36 & 0.25 & 0.01 \\ \hline
RP+$K$-means & 0.51  & 0.51 & 0.26 &  0.63 &  0.66 &  0.53 & 0.60  & 0.50 & 0.43 \\ \hline
DBSCAN & 0.63  & 0.62  & 0.43  & 0.71 & 0.71 &  0.63 & 0.36 & 0.25 & 0.01 \\ \hline
SC & 0.54 & 0.45  & 0.24 & 0.83 & 0.88 & 0.80 & 0.62  & 0.52  & 0.44\\ \hline
GMM & 0.44 & 0.35  & 0.02 &  0.64 & 0.61 & 0.55 & 0.42 & 0.31 & 0.10 \\ \hline
SMCE & 0.52 & 0.45  & 0.22 & 0.47 & 0.42 &  0.30  & 0.36 & 0.26 & 0.01 \\ \hline
HNMF & 0.41 & 0.32  & -0.02 & 0.63 & 0.66 & 0.53   & 0.36 & 0.25 & 0.00 \\ \hline
FMS & 0.57 & 0.50  & 0.27 & 0.70 & 0.81 & 0.65 & 0.74 & 0.70 & 0.65 \\ \hline
FSFDPC  & 0.58 & 0.51 &  0.26 & 0.63 &  0.61 & 0.54  &  0.36 & 0.25 & 0.00 \\ \hline
DL & 0.67 & 0.62 & 0.44 & 0.83 &  0.88  & 0.79 & 0.81 & 0.72 & 0.74 \\ \hline
DLSS & \underline{0.85}  & \underline{0.82}  & \underline{0.75} & \underline{0.85} & \underline{0.90} & \underline{0.81} & \underline{0.83} & \underline{0.73}  & \underline{0.76} \\ \hline
SRDL & \textbf{0.89}  & \textbf{0.92}  & \textbf{0.83} & \textbf{0.90} & \textbf{0.93} & \textbf{0.87} & \textbf{0.85} & \textbf{0.75}  & \textbf{0.79} \\ \hline
\end{tabular}
\end{adjustbox}
\caption{\label{tab:Summary}Summary of quantitative analyses of real HSI clustering; best results are in bold, second best are underlined.  The datasets have been abbreviated as I.P. (Indian Pines), S.A. (Salinas A), and K.S.C. (Kennedy Space Center).  In all cases, the proposed method offers the strongest overall performance, and in particular outperforms the DLSS algorithm, which does not regularize the underlying diffusion process.}
\end{table}

\subsection{Parameter Analysis}

\begin{figure}[htb!]
\centering
\begin{subfigure}{.15\textwidth}
\includegraphics[width=\textwidth]{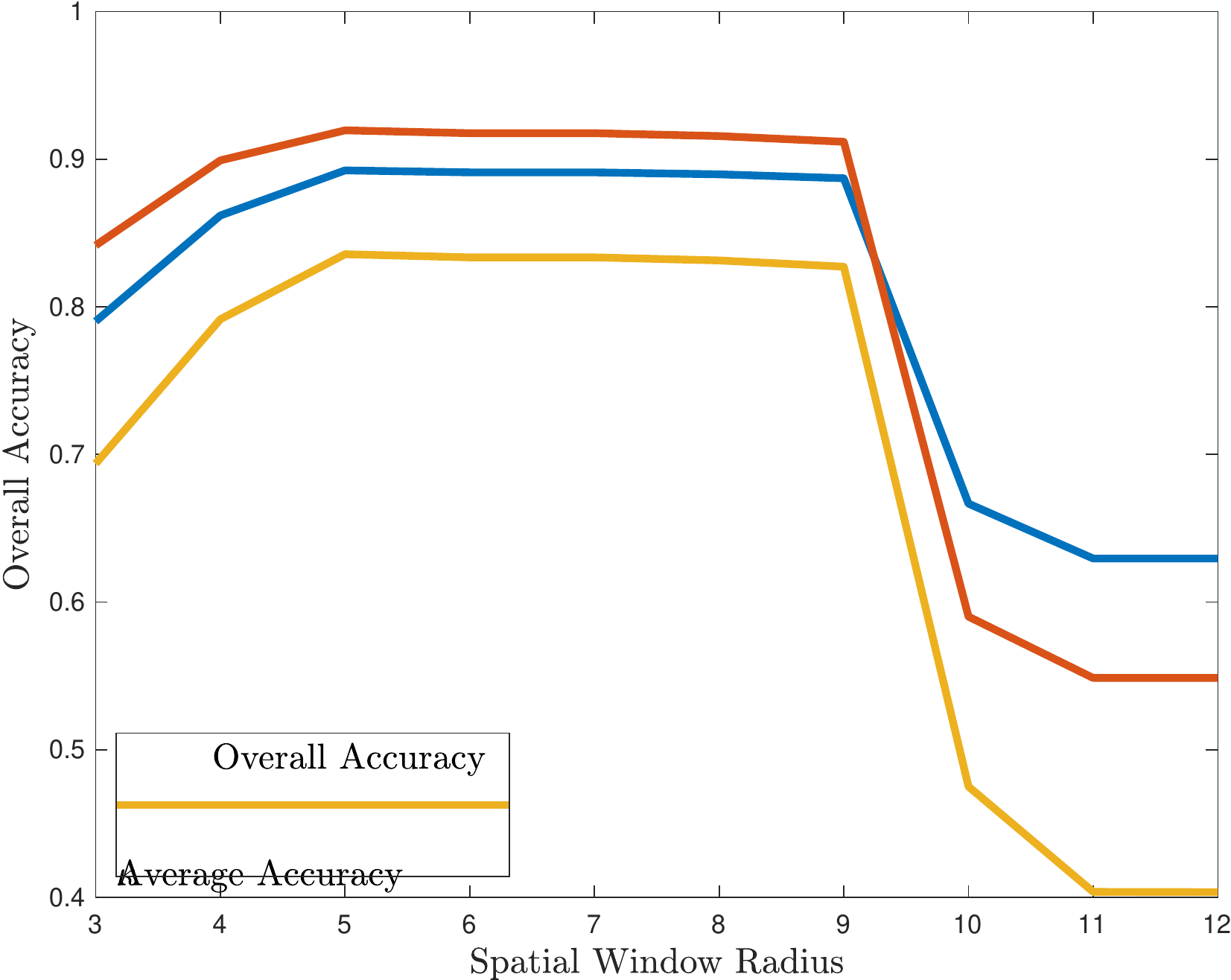}
\caption{IP}
\end{subfigure}
\begin{subfigure}{.15\textwidth}
\includegraphics[width=\textwidth]{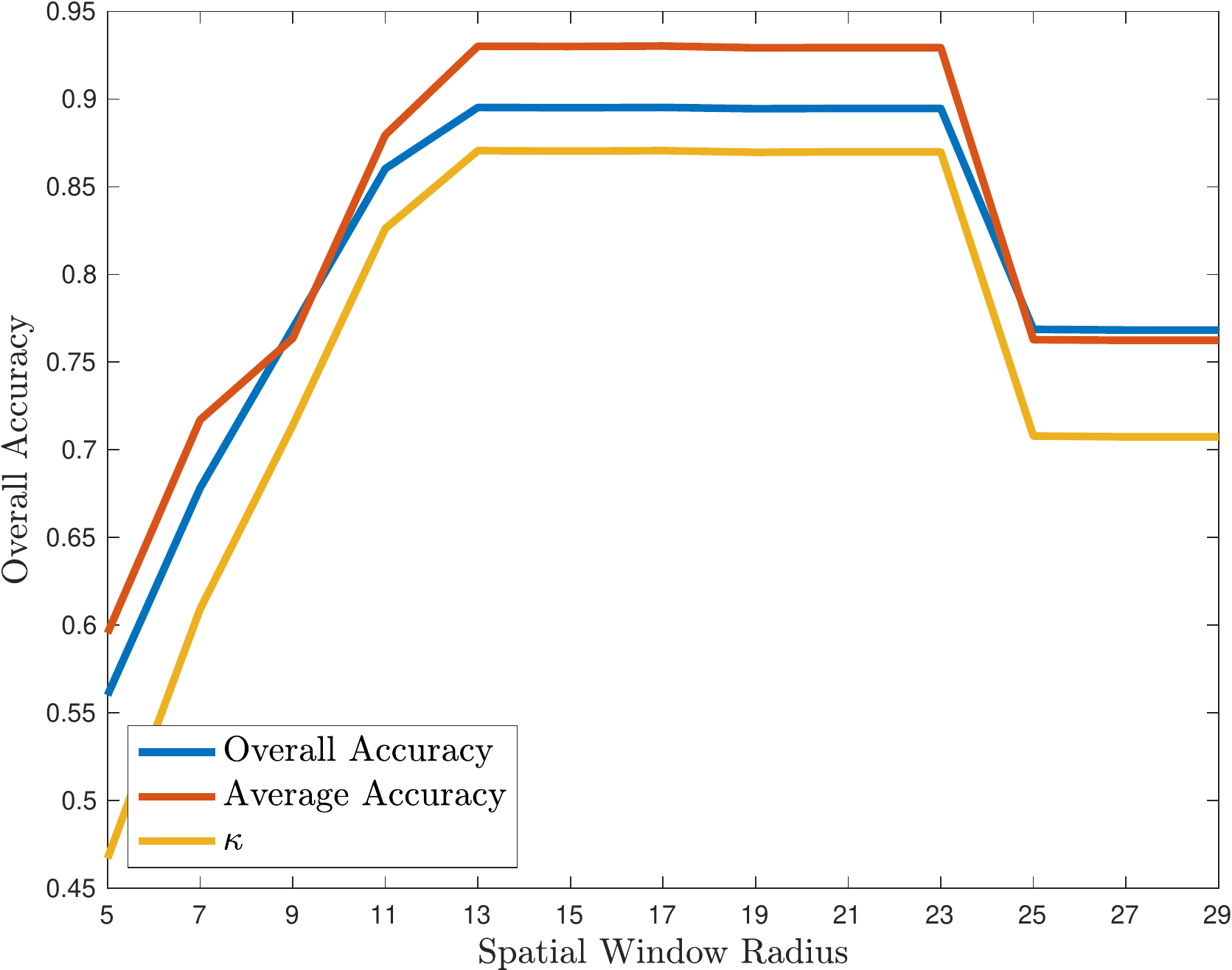}
\caption{SalinasA}
\end{subfigure}
\begin{subfigure}{.15\textwidth}
\includegraphics[width=\textwidth]{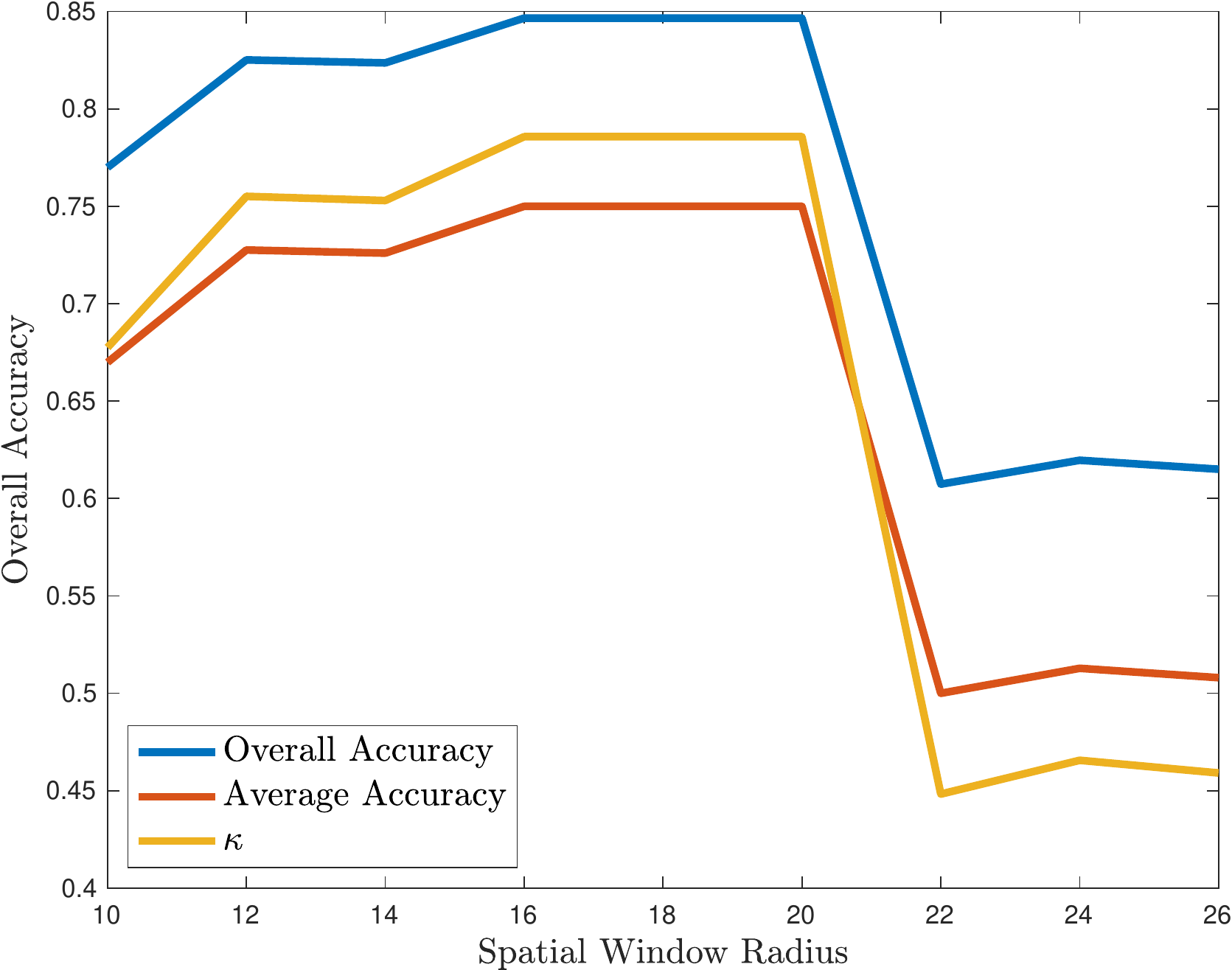}
\caption{KSC}
\end{subfigure}
\caption{\label{fig:SpatialRadius}Impact of spatial radius parameter.  Number of nearest neighbors is fixed at 100.  As the spatial window increases, empirical results improve, before decreasing.  This illustrates that optimal spatial regularization in the construction of diffusion maps is to consider a not too small (which leads to underregularization) and not too large (which leads to overregularization) spatial radius.  We note that once a good radius has been found, results are relatively robust.} 
\end{figure}

The crucial parameter in the proposed method is the spatial radius $r$ which determines how near the nearest neighbors in the underlying diffusion process must be.  The impact of this parameter in terms of overall accuracy, average accuracy, and $\kappa$ appear in Figure \ref{fig:SpatialRadius}.  The plots exhibit the tradeoff typical of regularization in machine learning: insufficient or excessive regularization are both detrimental.  It is critical to find a good range of regularization parameters, and the flat regions near the maxima in Figure \ref{fig:SpatialRadius} suggest the proposed method is relatively robust to the choice of $r$.  We remark that if knowledge of the spatial smoothness of the image is known or can be estimated a priori, then $r$ can be estimated.  

\section{Conclusions and Future Directions}\label{sec:Conclusions}

The incorporation of spatial regularity into the construction of the the underlying diffusion geometry improves empirical performance of diffusion learning for HSI clustering in all three datasets considered.  For images whose labels are sufficiently smooth, our results suggest there will be a regime for choices of spatial window in which incorporating spatial proximity improves the underlying mode detection and consequent labeling of HSI.  

In terms of \emph{computational complexity}, it suffices to note that the bottleneck is in the construction of $P$.  Since $k$ nearest neighbors are sought, and neighbors are constrained to live within a spatial radius $r$, as long as $r, k=O(1)$ with respect to $n$, the nearest neighbor searches for all points can be done in $O(n)$.  This gives an overall complexity for the algorithm that is essentially linear in $n$.  When the spatial radius is is large enough so that the full scene is considered in the nearest neighbor search, indexing structures (e.g. cover trees) allow for fast nearest neighbor searches, giving a quasilinear algorithm.

The proposed method is likely to be of substantial benefit for scenes that are in some sense smooth with respect to the underlying class labels.  For images with many classes that are rapidly varying in space---for example, urban HSI---alternative approaches for incorporating spatial features may be necessary.  One approach would be to consider as underlying data points not individual pixels, but higher order semantic features, for example image patches \cite{Buades2005_Non}.  This would allow for information about fine features such as edges and textures to be incorporated into the diffusion process, allowing for fine-scale clusters to be learned.  On the other hand, using patches instead of distinct pixels as the underlying dataset will increase the dimensionality of the data from $D$ to $D\cdot \text{(Patch Size)}$, which may lead to slower computational time.  However, it is expected that the patch space is low-dimensional \cite{Szlam2008regularization}, making the statistical learning problem tractable.  

\section*{Acknowledgments}
This research was partially supported by NSF-ATD-1737984, AFOSR FA9550-17-1-0280, and NSF-IIS-1546392.

\bibliographystyle{unsrt}
\bibliography{GRSL.bib}

\end{document}